\documentclass[lettersize,journal]{IEEEtran}
\usepackage{amsmath,amsfonts}

\usepackage{algorithm}
\usepackage{algpseudocode}
\usepackage{array}
\usepackage[caption=false,font=normalsize,labelfont=sf,textfont=sf]{subfig}
\usepackage{textcomp}
\usepackage{stfloats}
\usepackage{url}
\usepackage{verbatim}
\usepackage{graphicx}
\usepackage{cite}
\usepackage{amssymb}
\usepackage{hyperref}
\usepackage{wrapfig}
\usepackage{booktabs}
\usepackage{soul}
\usepackage{color}
\usepackage{amsthm}
\usepackage{subcaption}
\usepackage{mathtools}
\usepackage{multirow}
\usepackage{booktabs}
\usepackage{bbding}
\usepackage{cuted}   
\usepackage{capt-of}  
\usepackage{tabularx}

\begin{document}

\title{PRIMAL3: Pathfinding via Reinforcement and Imitation \\
Multi-Agent Learning - Leveraging LaCAM3
\\ {\large Project Page: \url{https://marmotlab.github.io/PRIMAL3/}}}

\author{Chengyang He$^{1}$, Tanishq Duhan$^{1}$, Gadiel Sznaier Camps$^{2}$, Fangyuan Wang$^{3}$, Yuhong Cao$^{1}$, \\ Jiankai Sun$^{2}$, Ge Sun$^{1}$, Mac Schwager$^{2}$, Guillaume Sartoretti$^{1,\dagger}$
\vspace{-5em}
\thanks{$\dagger$ Corresponding author: guillaume.sartoretti@nus.edu.sg}
\thanks{$^1$ Multi-Agent Robotic Motion Lab (MARMot), National University of Singapore.}
\thanks{$^2$ Multi-robot Systems Lab (MSL), Stanford University.}
\thanks{$^3$ Robotics and Machine Intelligence Lab (ROMI), Hong Kong Polytechnic University. He contributed to this work while visiting the MARMoT Lab, National University of Singapore.}}

\markboth{Journal of \LaTeX\ Class Files,~Vol.~*, No.~*, July~2026}%
{Shell \MakeLowercase{\textit{et al.}}: A Sample Article Using IEEEtran.cls for IEEE Journals}

\maketitle
\vspace{-5em}
\begin{strip}
    \centering
    \includegraphics[width=\textwidth]{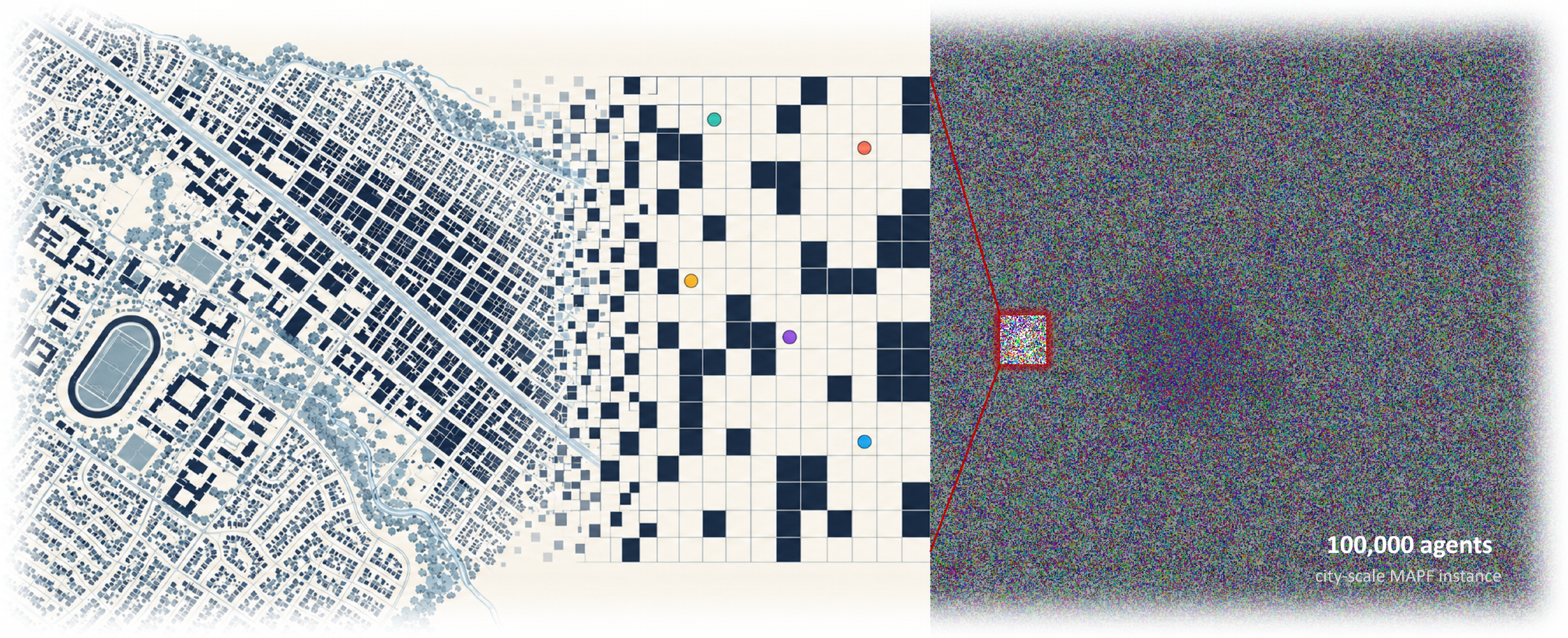}
    \captionof{figure}{From urban mobility to ultra-large-scale MAPF.
        Mobile entities and inaccessible city blocks can conceptually be abstracted as agents and obstacles in a discrete MAPF environment (left). 
        PRIMAL3 coordinates city-scale number of agents (100,000 in practice) simultaneously in the large random map (right).}
    \label{fig:banner_long}
\end{strip}
\vspace{-1em}

\begin{abstract}

We present PRIMAL3, an ultra-large-scale learning-based framework for multi-agent pathfinding (MAPF) that integrates reinforcement learning, topology-aware communication, LaCAM3-guided training, and PIBT-based action refinement. 
PRIMAL3 targets failures at topologically critical states, where agents must coordinate decisively around bottlenecks, dead ends, and persistent conflicts. 
Each agent is represented using features derived from cut vertices, dead-end regions, shortest-path distances, and blocking estimates. 
Two complementary graphs capture agent interactions: a same-direction following graph propagates multihop context along compatible paths, while a different-direction conflict graph differentiates agents competing for shared space through masked attention and relative features.
During training, we propose to let policy entropy identify uncertain agents, for which LaCAM3 provides confidence-triggered action interventions and label-smoothed imitation targets. 
During execution, a priority-aware PIBT module refines the proposed joint actions using persistent, learned, and distance-aware priorities together with policy-aware fallback preferences while maintaining collision-free execution. 
The resulting framework combines learned exploration with structured expert guidance without requiring LaCAM3 at inference.
Experiments demonstrate that PRIMAL3 substantially outperforms state-of-the-art learning-based baselines and scales to ultra-large instances with up to city-level 100,000 agents.
Real-world experiments further demonstrate the feasibility of deploying PRIMAL3 on physical robotic systems and ablation studies validate the individual contributions the components we proposed.

\end{abstract}


\section{Introduction}
\label{sec:intro}

\IEEEPARstart{M}{ulti-agent} path finding (MAPF) is a fundamental problem in multi-robot coordination, with applications in automated warehouses~\cite{li2021lifelong,wang2020mobile}, traffic management~\cite{duhan2025p3gasus}, intersection coordination~\cite{li2023intersection}, and other multi-agent systems~\cite{stern2019multi,ma2017feasibility}. 
In classical one-shot MAPF, each agent is assigned a start and a goal vertex on a graph, and the objective is to compute collision-free paths for all agents. 
Despite this simple formulation, MAPF is computationally challenging because the feasible motion of one agent depends on the decisions of many others.
This coupling becomes particularly pronounced in dense environments containing narrow corridors, bottlenecks, and other strong topological constraints.

Search-based MAPF research has produced optimal~\cite{wagner2011m,sharon2015conflict}, bounded-suboptimal~\cite{wagner2015subdimensional,barer2014suboptimal,li2021eecbs}, and anytime suboptimal~\cite{li2021anytime} solvers. 
Conflict-Based Search (CBS)~\cite{sharon2015conflict} and its variants provide strong theoretical guarantees, while scalable methods such as MAPF-LNS~\cite{li2021anytime,li2022mapf} and LaCAM~\cite{okumura2023lacam,okumura2023engineering} achieve excellent empirical performance on large instances. 
Nevertheless, these methods generally compute centralized joint plans, and deviations from the planned state may require costly replanning. 
In robotic systems subject to environmental changes, execution noise, or incomplete observations, policies that can react directly to the current state are therefore attractive.

Learning-based MAPF treats the planning process into a policy that can generalize across maps and agent configurations. 
Early methods such as PRIMAL~\cite{sartoretti2019primal} combine reinforcement learning (RL) with imitation learning (IL) to learn decentralized pathfinding policies from both environmental interaction and expert demonstrations. 
Subsequent approaches have introduced richer observations~\cite{wang2020mobile,liu2020mapper}, communication mechanisms~\cite{li2020graph,li2021message,jain2026pairwise}, graph representations~\cite{he2024alpha}, and more expressive policy architectures~\cite{andreychuk2025mapf,andreychuk2025advancing}. 
Despite this progress, learning-based MAPF still faces three closely related challenges:

\textit{1) Critical MAPF interactions are topologically structured and relationally heterogeneous.}
Local occupancy observations alone do not explicitly reveal whether an agent is approaching a cut vertex, entering a dead-end region, or occupying a position that blocks access between different parts of the map. 
Previous framework also uses reference-path overlap to identify agents that are likely to interact and should therefore exchange information~\cite{he2025social}.
Although this heuristic provides a useful estimate of interaction relevance, path overlap alone does not distinguish the type of coordination required. 
Agents moving along compatible, same-direction paths benefit from sharing route-level context and maintaining coherent motion, whereas agents approaching the same constrained region from conflicting directions must differentiate their decisions by proceeding, yielding, waiting, or selecting alternative motions. 
In other words, path-overlap-based selection determines \emph{who} should communicate but does not specify \emph{how} information from different types of interactions should be processed. 
Existing homogeneous aggregation can blur these relational differences, particularly under parameter sharing, where all agents employ the same policy structure. 
Therefore, we propose that effective communication should account for both the underlying map topology and the heterogeneous relations between interacting agents.

\textit{2) Policy uncertainty is most costly when coordination requires commitment.}
In open regions, several actions may be similarly effective, and a high-entropy policy may have little consequence. 
Near a bottleneck or dead end, however, successful coordination may depend on establishing a clear right-of-way or committing to a specific avoidance behavior. 
Episodic RL objectives only provide indirect supervision for such decisions, and the learned policy may continue to assign comparable probabilities to several locally plausible actions. 
Repeated sampling from this multimodal distribution can produce inconsistent decisions across successive timesteps, leading to hesitation, blocking, or oscillatory behavior. 
This motivates targeted expert guidance that sharpens the policy specifically when its action preference remains uncertain.

\textit{3) Collision-free action refinement does not necessarily ensure coordinated progress.}
Because decentralized policies select actions independently, their proposals may contain vertex or edge conflicts. 
Execution-time shielding addresses this issue by refining the proposed actions into a collision-free joint action before they are executed. 
In particular, PIBT-based shields assign actions according to agent priorities and resolve conflicts through priority inheritance and backtracking~\cite{veerapaneni2024improving,jiang2025deploying,jain2026pairwise}. 
However, this refinement primarily guarantees one-step collision avoidance rather than coordinated progress over time. 
When PIBT is reinitialized independently at every timestep, its priorities do not retain information about previous delays, allowing some agents to be repeatedly blocked. 
Moreover, if the shield receives only a discrete action from each policy, the policy’s relative preferences over alternative actions are discarded. 
Consequently, under repeated contention, fallback decisions may depend on transient priorities or arbitrary tie-breaking, even though each executed joint action remains collision-free. 
Practical shielding should therefore maintain priority information across timesteps and preserve the learned policy’s fallback preferences.

Motivated by these observations, we propose PRIMAL3, a topology-aware learning framework that integrates structured graph communication, LaCAM3-guided training, and priority-aware PIBT action refinement. 
Given the current agent positions and goals, PRIMAL3 first computes individual A*-based reference paths as compact estimates of the agents' navigation intentions.
A following and conflict graph are generated using directional path overlaps to provide agents with information about compatible and conflicting paths. 
A sparse neighbor-selection mechanism converts these interaction graphs into communication graphs that retain the most critical and relevant relationships. 
Each agent is represented using topology-aware nodes, by which there are hand-crafted features to directly expose bottlenecks, constrained regions and reachability dependencies. 
These features provide necessary structural information that complement the agent's learned latent observation representation.
During the early portion of a expert-involved training episode, PRIMAL3 uses LaCAM3 to provide selective action interventions for identified decision uncertainty via entropy of each agent's action distribution.
Later portion of this episode's conflicts are resolved using expert guidance, incorporated as part of a categorical cross-entropy loss. 
This confidence-guided supervision encourages decisive, coordinated actions while retaining the exploratory benefits of RL. 
Finally, to support safe and temporally consistent execution, PRIMAL3 employs a priority-aware PIBT-based action refinement module. 
It combines persistent waiting-time information with learned coordination priorities and goal distance to resolve contention, while preserving the policy’s preferences when selecting alternative actions. 
This refinement retains PIBT’s collision-avoidance capability while reducing repeated blocking and arbitrary fallback decisions in dense environments.

Extensive evaluations on random and maze maps demonstrate that PRIMAL3 consistently outperforms existing learning-based MAPF baselines and scales to ultra-large instances with up to \textbf{city-level 100,000} agents (as shown in Fig~\ref{fig:banner_long}). 
Notably, on random maps, PRIMAL3 achieves success rates comparable to those of state-of-the-art search-based solvers, including LNS2 and LaCAM3, substantially narrowing the long-standing performance gap between learning-based and classical MAPF methods. 
Ablation studies confirm the individual contributions of our proposed components, while real-world multi-robot experiments demonstrate the feasibility of deploying PRIMAL3 on physical robotic systems.

Our contributions are summarized as follows:
\begin{itemize} 
    \item \textbf{Dual-graph Representation}: We introduce a topology-aware dual-graph representation for learning-based MAPF. Explicit structural node features are combined with separate following and conflict graphs to represent path-compatible and path-competing interactions.

    \item \textbf{Confidence-boosting Expert Intervention}: We introduce a LaCAM3-guided confidence-boosting strategy for policy training. Entropy-based uncertainty detection triggers selective expert intervention, while label-smoothed expert targets provide an additional imitation-learning objective. This strategy encourages more decisive and coordinated actions in challenging states.

    \item \textbf{PIBT-based Action Refinement}: We propose a priority-aware PIBT action refinement module that combines persistent, learned, and distance-aware agent priorities with policy-aware fallback preferences, improving collision-free execution and reducing arbitrary action repair in dense environments.
\end{itemize}

\section{Prior Work}

\subsection{Traditional Methods}

Research on Multi-Agent Path Finding (MAPF) originated from classical planning methods, which remain the algorithmic foundation of the field~\cite{stern2019multi}. 
Traditional MAPF solvers are commonly categorized into three groups according to their solution guarantees: optimal, bounded-suboptimal, and unbounded-suboptimal methods~\cite{gao2023review}. 
This taxonomy also reflects the historical evolution of the field, from exact but computationally demanding solvers toward increasingly scalable methods designed for large teams and real-time deployment.

Optimal MAPF methods aim to compute collision-free solutions with minimum total cost. 
Among the most influential representatives are M*~\cite{wagner2011m} and Conflict-Based Search (CBS)~\cite{sharon2015conflict}. 
M* performs coupled search only when conflicts arise, thereby avoiding full joint-space expansion unless necessary. 
CBS adopts a two-level architecture in which the high level resolves conflicts over a constraint tree, while the low level computes individually optimal paths, typically using A*, under the imposed constraints. 
Owing to its conceptual simplicity and modularity, CBS has become one of the most important backbones in MAPF and has inspired a large family of subsequent methods. 

A substantial line of work has focused on improving the efficiency of CBS without changing its basic structure. 
Early extensions such as ICBS~\cite{boyarski2015icbs} improve the high-level search through conflict prioritization and bypassing, while later methods introduce increasingly strong admissible heuristics~\cite{felner2018adding,li2019improved}. 
Another important direction is symmetry reasoning, including rectangle reasoning, corridor reasoning, target reasoning, and mutex propagation, all of which reduce repeated exploration caused by recurring conflict patterns~\cite{li2019symmetry,li2020new,zhang2022multi,li2021pairwise}. 
In parallel, disjoint splitting~\cite{li2019disjoint} improves the branching strategy itself by generating disjoint subproblems and thus reducing duplicated search effort. 
These advances have made CBS-based solvers substantially more competitive in practice, and they remain central to the design of exact and bounded-suboptimal MAPF planners. 
To improve scalability beyond exact search, bounded-suboptimal planners trade strict optimality for computational efficiency through explicit approximation guarantees. Inflated M*~\cite{wagner2015subdimensional} relaxes M* through heuristic inflation, while ECBS~\cite{barer2014suboptimal} accelerates CBS by replacing both levels of search with focal search. EECBS~\cite{li2021eecbs} further strengthens this line by combining explicit estimation search at the high level with focal search at the low level, achieving strong empirical performance while preserving bounded-suboptimal guarantees. Together, these methods represent an important middle ground between exact optimality and practical efficiency.
Recent closed-loop CBS work~\cite{li2026adaptive} also explicitly positions ICBS, ECBS, and EECBS as key milestones within this broader CBS lineage.

As MAPF has moved toward large-scale warehouse and logistics applications, the focus of the community has gradually shifted from strict guarantees to solver effectiveness. 
This trend has driven the development of unbounded-suboptimal planners, whose primary goals are high success rate, low runtime, and strong empirical scalability. 
One influential line is based on large neighborhood search (LNS). 
MAPF-LNS~\cite{li2021anytime} first computes an initial solution using any efficient MAPF solver and then repeatedly replans selected subsets of agents to improve solution quality. 
MAPF-LNS2~\cite{li2022mapf} further develops this idea by first allowing collisions in the initial solution and then iteratively repairing conflicting subsets under a limited time budget. 
These methods demonstrate that, for large instances, iterative repair can be substantially more scalable than directly solving the full problem to high quality from scratch.

Another highly influential line is prioritized and configuration-based planning. 
Prioritized planning~\cite{silver2005cooperative} has long played an important role in MAPF, and more recent methods such as PIBT~\cite{okumura2022priority} and LaCAM~\cite{okumura2023lacam} have pushed this direction significantly further. 
PIBT is an online priority-based method that generates one action per agent at each timestep through priority inheritance and backtracking.
Building on this idea, LaCAM introduces a two-level configuration-based search framework, where the high level searches over joint configurations of all agents and the low level rapidly generates successor configurations. 
This design yields strong scalability while maintaining high solution quality. 
More recent developments, including LaCAM* and the engineering-oriented version commonly referred to as LaCAM3~\cite{okumura2023engineering}, have made this family one of the most competitive directions in traditional MAPF. 
In fact, very recent work~\cite{li2025fico} still describes LaCAM* as the current state of the art among anytime configuration-based solvers, and proposes additional guidance mechanisms specifically to further improve LaCAM*-style search rather than replace it.

Recent traditional MAPF research has also revisited the planning paradigm itself. 
Classical solvers such as CBS, ECBS, EECBS, and LaCAM are open-loop, computing a full plan before execution. 
To better handle disturbances and online changes, recent work has introduced finite-horizon and closed-loop MAPF formulations, including Finite-Horizon Hierarchical Factorization~\cite{li2025multi}, FICO~\cite{li2025fico}, and ACCBS~\cite{li2026adaptive}.

Overall, the development of traditional MAPF methods reflects a clear progression from exact global planning to scalable and deployment-oriented coordination. 
Optimal and bounded-suboptimal solvers, especially the CBS family, remain indispensable when formal guarantees are required. 
At the same time, unbounded-suboptimal methods, particularly the LaCAM family, have become increasingly important because they offer an exceptional balance among runtime, scalability, and solution quality. 
Meanwhile, emerging finite-horizon and closed-loop methods such as FICO and ACCBS expand traditional MAPF toward more realistic execution settings. 
Nevertheless, for the standard one-shot MAPF problem, LaCAM-based solvers remain among the most important and competitive baselines, and continue to define the practical performance frontier of traditional MAPF.

\subsection{Learning-based Methods}

Learning-based MAPF methods can be broadly divided into two main lines: reinforcement learning (RL)-based approaches and imitation learning (IL)-based approaches. 
Although many frameworks combine both, one paradigm usually plays the primary role while the other serves as auxiliary supervision or fine-tuning. 
Early work in learning-based MAPF can be traced back to PRIMAL~\cite{sartoretti2019primal}, a hybrid RL-IL framework in which imitation learning serves as an auxiliary means to enhance agents' ability to find better solutions during the exploration phase. 
Building on this foundation, subsequent work increasingly focused on strengthening the RL component through improved training algorithms and more informative heuristic observations. 
For instance, G2RL~\cite{wang2020mobile} augments agent observations with A* paths and encourages agents to follow them strictly through reward design. 
MAPPER~\cite{liu2020mapper} relaxes this strict reliance on expert paths by introducing a more flexible reward structure. 
DHC~\cite{ma2021distributed} further reduces this dependency by replacing A*-based guidance with heuristic observations derived from Breadth-First Search (BFS), while ALPHA~\cite{he2024alpha} expands the observation space through graph-based representations, enabling agents to consider a broader range of suboptimal alternatives yet potentially beneficial actions.

Beyond improving individual observation and representation design, another important direction has been to enhance coordination through explicit or implicit communication. 
Several RL-based MAPF frameworks with inner-team communication have been proposed along this line. 
PICO~\cite{li2022multi} builds cluster-based communication through implicit prioritization; 
SCRIMP~\cite{wang2023scrimp} adopts transformer-based communication learning; 
DCC~\cite{ma2021learning}, an extension of DHC, identifies more relevant communication partners through decision causal units; 
SIGMA~\cite{liao2025sigma} avoids explicit message passing at execution time, but learns a consensus-building mechanism through inter-agent information exchange during training; 
and SYLPH~\cite{he2025social} constructs a more efficient communication topology based on conflicts among agents’ optimal paths, allowing agents to exchange social preferences selectively. 
Despite their architectural differences, these methods share a common goal: improving coordination by encoding local observations into richer representations and learning which agents should exchange information.

In recent years, advances in large-language models (LLMs), vision-language-action (VLA) systems, and related techniques have significantly strengthened imitation learning (IL) as a general paradigm for sequential decision-making. 
As a result, IL-centered frameworks have attracted increasing attention in the MAPF community. 
Existing IL-based MAPF methods can be roughly grouped into three categories.
The first category consists of conventional behavior cloning (BC) approaches. 
The work in~\cite{li2020graph} is, to the best of our knowledge, the first to adopt pure imitation learning as the primary training paradigm for MAPF. 
By combining BC with graph neural networks, it enables local communication among agents and establishes a fully IL-based framework. 
Building on this line of work, MAGAT~\cite{li2021message} further improves communication efficiency by using graph attention mechanisms to filter and weight local neighbors more selectively.
The second category moves beyond standard imitation learning and incorporates distributed expert algorithms as a post-processing collision-shielding module~\cite{virmani2021subdimensional,morag2023adapting,veerapaneni2024improving,veerapaneni2025work}. 
Unlike earlier methods~\cite{sartoretti2019primal,wang2023scrimp,he2025social} that use post-processing only for reactive collision avoidance, these approaches not only prevent collisions but also continue to guide agents toward their goals during execution.
This leads to a stronger and more effective form of collision shielding, improving both safety and task progress.
The third category includes foundation-model-scale frameworks~\cite{andreychuk2025advancing}. 
MAPF-GPT~\cite{andreychuk2025mapf} is the first general MAPF framework designed to handle diverse map types using a large-scale model trained on extensive datasets, with a network size of 85 million parameters. 
More recently, HMAGAT~\cite{jain2026pairwise} integrates the strengths of the three IL-based paradigms discussed above into a unified framework. 
By combining efficient local communication, expert-guided shielding, and large-scale learning capacity, it achieves performance beyond previous approaches and represents the current state of the art among learning-based MAPF methods.

\section{Problem Statement}

\subsection{Definition of One-shot Multi-Agent Pathfinding}

The multi-agent pathfinding (MAPF) problem has been studied under a variety of settings, including the classical one-shot formulation, lifelong MAPF, MAPF with kinematic constraints, prioritized MAPF, and multi-agent pickup-and-delivery variants~\cite{li2022mapf,ma2019lifelong,skrynnik2023learn,damani2021primal,chandra2023socialmapf,okumura2022priority,zang2025online,zheng2026learning,zhang2026optimization,shaoul2024multi,shaoul2025collaborative,zhang2025flow,ma2017lifelong}. In this work, we focus on the classical one-shot MAPF setting.

A one-shot MAPF instance is defined on an undirected graph $\mathcal{G}=(\mathcal{V},\mathcal{E})$, where $\mathcal{V}$ denotes the set of traversable vertices and $\mathcal{E}$ denotes the set of edges connecting neighboring vertices. 
An edge $(u,v)\in\mathcal{E}$ indicates that an agent can move from vertex $u$ to vertex $v$ in one discrete timestep. 
Let $\mathcal{A}=\{a_1,\dots,a_n\}$ denote the set of agents. Each agent $a_i$ is associated with a start vertex $s_i\in\mathcal{V}$ and a goal vertex $g_i\in\mathcal{V}$. 
The corresponding start and goal sets are denoted by $\mathcal{S}=\{s_1,\dots,s_n\}$ and $\mathcal{D}=\{g_1,\dots,g_n\}$, respectively.
One instance/configuration in MAPF context can be explained as the set of $\{\mathcal{G,S,D} \}$.

Time is modeled as discrete, i.e., $t\in\mathbb{N}$. 
At each timestep, an agent may either remain at its current vertex or move to one of its adjacent vertices. 
Accordingly, a path for agent $a_i$ can be written as
\begin{equation}
\begin{aligned}
    \tau_i = \left(\tau_i(0), \tau_i(1), \dots, \tau_i(T)\right),
\end{aligned}
\end{equation}
where $\tau_i(t)\in\mathcal{V}$ denotes the location of agent $a_i$ at time $t$, and $T$ is the planning horizon. 
A valid MAPF solution is a set of paths $\{\tau_i\}_{i=1}^{n}$ satisfying the following conditions:
\begin{equation}
\begin{aligned}
&\tau_i(0)=s_i,~\tau_i(T)=g_i,~\forall i\in\{1,\dots,n\},\\
&\tau_i(t)\neq\tau_j(t),~\forall i\neq j,~\forall t\in\{0,\dots,T\},\\
&\left(\tau_i(t),\tau_i(t+1)\right)\neq\left(\tau_j(t+1),\tau_j(t)\right),\forall i\neq j,\\
&\forall t\in\{0,\dots,T-1\}.
\end{aligned}
\end{equation}
The first condition requires every agent to start from its assigned initial vertex and reach its designated goal vertex. 
The second condition excludes vertex collisions, meaning that no two agents can occupy the same vertex at the same time.
The third condition excludes edge collisions, meaning that two agents cannot traverse the same edge in opposite directions simultaneously. 
Under these constraints, the objective of classical one-shot MAPF is to compute collision-free paths for all agents while minimizing the overall completion time.

\subsection{Graph-Theoretic Preliminaries for Multi-Agent Learning}

Graph representations provide a natural way to formulate both the environment structure or inter-agent interactions in multi-agent learning. 
In MAPF, the workspace is already modeled as a graph $\mathcal{G}=(\mathcal{V},\mathcal{E})$, where vertices denote feasible locations and edges denote one-step transitions. 
Beyond this physical environment graph, it is often useful to introduce an \emph{interaction graph} over the agent set $\mathcal{A}=\{a_1,\dots,a_n\}$ to capture which agents should exchange information or coordinate decisions. 
Such graph-based formulations are widely used in learning-based MAPF because they provide an explicit relational structure while remaining compatible with decentralized execution~\cite{battaglia2018relational,li2020graph,he2025social}.

Formally, at time $t$, we define a dynamic interaction graph $\mathcal{G}^{\text{int}}_t=(\mathcal{A},\mathcal{E}^{\text{int}}_t)$, where each node corresponds to an agent and each edge $(a_i,a_j)\in\mathcal{E}^{\text{int}}_t$ indicates that agent $a_j$ is relevant to agent $a_i$ at time $t$. 
The edge set can be constructed in different ways depending on the method, for example based on spatial proximity, local field of view overlap, predicted path conflicts, learned communication priorities, or even a complete graph~\cite{ma2021distributed,li2022multi,he2025social,li2021message,he2024alpha}. 
Each node may be associated with a feature vector $x_i^t$ encoding the local observation, goal information, or internal hidden state of agent $a_i$, while each edge may optionally carry a feature $e_{ij}^t$ describing pairwise relations~\cite{jain2026pairwise} such as relative position or conflict risk.

Given $\mathcal{G}^{\text{int}}_t$, graph-based multi-agent learning typically updates agent representations through message passing over neighboring nodes. 
In a generic form, the hidden state of agent $a_i$ can be written as:
\begin{equation}
    \begin{aligned}
        h_i^{t+1} = \phi\!\left(h_i^t,\ \bigoplus_{a_j\in\mathcal{N}_t(a_i)} \psi(h_i^t,h_j^t,e_{ij}^t)\right),
    \end{aligned}
\end{equation}
where $\mathcal{N}_t(a_i)$ denotes the neighbors of agent $a_i$ in the interaction graph, $\psi(\cdot)$ is a message function, $\bigoplus$ is a permutation-invariant aggregation operator, and $\phi(\cdot)$ is an update function. 
This formulation covers a wide range of graph-based architectures used in MAPF, including graph convolution, attention-based neighborhood selection, and transformer-style communication~\cite{li2020graph,li2021message,wang2023scrimp}. 
In this work, graph theory is therefore used not only to describe the underlying MAPF environment, but also to formalize structured coordination and information exchange among agents.

\section{PRIMAL3}

In this section, we introduce three main contributions of PRIMAL3: 1) representation learning based on dual graphs, 2) imitation learning based on LaCAM3, and 3) our modified PIBT shielding for action refinement. 

\begin{figure*}[t]
    \centering
    \includegraphics[width=\textwidth]{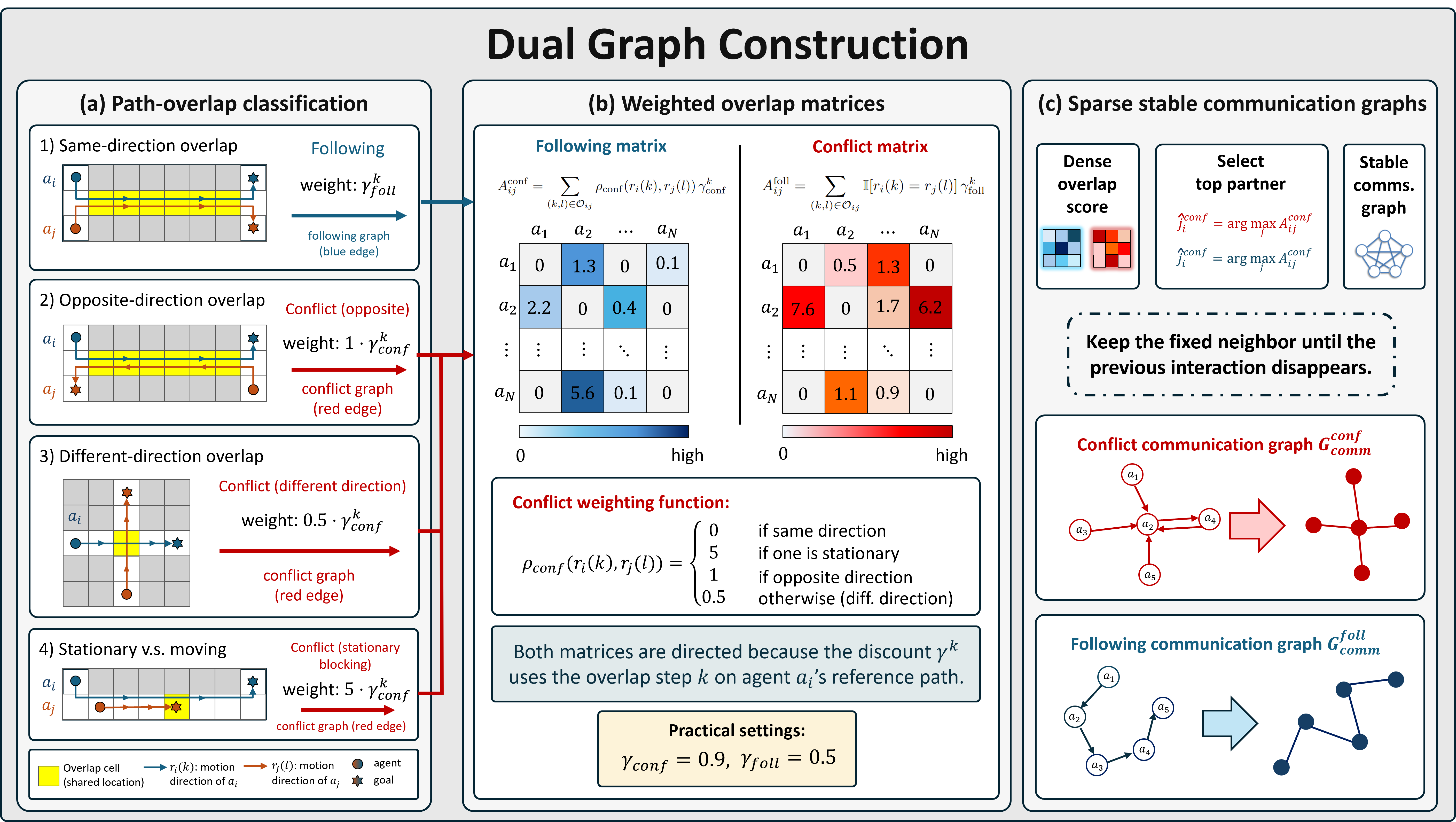}
    \caption{Dual-graph construction from reference-path overlaps. Same-direction overlaps are accumulated into a following graph, while opposite-direction, different-direction, and stationary-blocking overlaps are weighted into a conflict graph. The resulting dense overlap matrices are then used to select each agent’s most relevant conflict and following neighbors, forming sparse and temporally stable communication graphs.}
    \label{fig:dual_graph}
\end{figure*}

\subsection{Dual-Graph Representation Learning}

In MAPF, interactions among agents are shaped not only by spatial proximity but also by the underlying map topology, particularly in highly structured environments such as mazes. 
A*-computed reference paths capture the task-relevant map topology and provide a useful basis for identifying potential interactions among agents~\cite{he2025social}.
Based on their reference paths, nearby agents may require different forms of coordination: their paths may compete for the same constrained space or remain compatible by following a shared direction. 
To distinguish these interactions, we construct two complementary agent-interaction graphs: a conflict graph and a following graph. 
The conflict graph captures competitive interactions arising from incompatible paths, whereas the following graph represents compatible, directionally aligned movements.

\subsubsection{Reference Path and Direction Encoding}

Given the current map $\mathcal{G}$, agent positions $\mathcal{A}$ or $\mathcal{S}$, and goal locations $\mathcal{D}$, we first compute an individual reference path for each agent using an A* planner. Let
\begin{equation}
    \begin{aligned}
        p_i = \left(p_i(0), p_i(1), \dots, p_i(L_i)\right)
    \end{aligned}
\end{equation}
denote the reference path of agent $a_i$, where $p_i(k)$ is the grid location of agent $a_i$ at path step $k$. 
To evaluate pairwise path interactions over a common horizon, all reference paths are padded to the same length by repeating the corresponding goal locations.

For each reference path, we further assign a discrete motion direction to every transition according to the definition of cardinal directions for MAPF. 
The direction of agent $a_i$ at step $k$ is denoted by $r_i(k)$, where
\begin{equation}
    \begin{aligned}
        r_i(k)\in\{0,1,2,3,4\}
    \end{aligned}
\end{equation}
corresponds to staying still, moving right, moving up, moving left, and moving down, respectively. 
The terminal or padded portion of a path is assigned the movement \emph{stay still} $0$. 
These path-direction sequences provide the basis for distinguishing different types of pairwise path overlaps.

\subsubsection{Conflict and Following Graph Construction}

Based on the reference paths and their direction sequences, we compute pairwise interaction scores between agents. 
If two agents visit the same grid location along their reference paths, their interaction is classified according to their motion directions at the overlapping location, as shown in Fig~\ref{fig:dual_graph}. 
Same-direction overlaps are assigned to the following graph, while opposite-direction and different-direction overlaps are assigned to the conflict graph. 
Opposite-direction overlaps are treated as stronger conflicts than non-opposite directional mismatches because the opposite-direction overlaps always happens in the narrow corridor while the non-opposite overlaps might also exist in the open space. 
It is worth noting that when one agent is stationary at an overlapping location while another agent moves through it, the interaction is also treated as a conflict, since this situation often corresponds to potential blocking in narrow passages or maze-like environments after one agent arrive its goal.

We define two directed weighted adjacency matrices which can be calculated by Algorithm~\ref{alg:dual_graph_construction},
\begin{equation}
    \begin{aligned}
        A^{\mathrm{conf}} \in \mathbb{R}^{n\times n},~
        A^{\mathrm{foll}} \in \mathbb{R}^{n\times n},
    \end{aligned}
\end{equation}
where $A^{\mathrm{conf}}_{ij}$ measures the conflict influence imposed on agent $a_i$ by agent $a_j$, and $A^{\mathrm{foll}}_{ij}$ measures the same-direction influence imposed on agent $a_i$ by agent $a_j$. 
Let $\mathcal{O}_{ij}=\{(k,l)\mid p_i(k)=p_j(l)\}$ denote the set of overlapping positions between the reference paths of agents $a_i$ and $a_j$. 
The directed conflict and same-direction adjacency matrices are defined as
\begin{equation}
    \begin{aligned}
        A^{\mathrm{conf}}_{ij}=
        \sum_{(k,l)\in\mathcal{O}_{ij}}
        \rho_{\mathrm{conf}}\!\left(r_i(k),r_j(l)\right)\gamma_{\mathrm{conf}}^{k},\\
        A^{\mathrm{foll}}_{ij}=
        \sum_{(k,l)\in\mathcal{O}_{ij}}
        \mathbb{I}\!\left[r_i(k)=r_j(l)\right]\gamma_{\mathrm{foll}}^{k},
    \end{aligned}
\end{equation}
where $r_i(k)$ is the direction of agent $a_i$ at step $k$, and $\gamma_{\mathrm{conf}}$ and $\gamma_{\mathrm{foll}}$ are temporal discount factors. The conflict coefficient is defined as
\begin{equation}
    \rho_{\mathrm{conf}}(r_i,r_j)=
        \begin{cases}
        0, & r_i=r_j,\\
        5, & r_i=0\ \text{or}\ r_j=0,\\
        1, & (r_i,r_j)\in\text{opposite directions},\\
        0.5, & \text{otherwise}.
        \end{cases}
\end{equation}
The adjacency matrices are directed because the same overlap may occur at different distances along the two agents' reference paths; 
therefore, $A^{\mathrm{conf}}_{ij}$ and $A^{\mathrm{foll}}_{ij}$ are discounted according to the overlap step on agent $a_i$'s path.
We set $\gamma_{\mathrm{conf}}=0.9$ and $\gamma_{\mathrm{foll}}=0.5$ in practice. The larger conflict discount preserves the influence of future conflicts, allowing agents to react early to potential blocking or head-on encounters. 
In contrast, same-direction overlaps mainly provide short-term consistency cues, so a smaller discount emphasizes immediate cooperation while suppressing distant overlaps.

\begin{algorithm}[t]
\caption{Construction of Conflict and Following Graphs}
\label{alg:dual_graph_construction}
\begin{algorithmic}[1]
\Require Reference paths $\{p_i\}_{i=1}^{n}$, direction sequences $\{r_i\}_{i=1}^{n}$, discounts $\gamma_{\mathrm{conf}}$, $\gamma_{\mathrm{foll}}$
\Ensure Directed adjacency matrices $A^{\mathrm{conf}}$ and $A^{\mathrm{foll}}$
\State Initialize $A^{\mathrm{conf}}\leftarrow \mathbf{0}_{n\times n}$ and $A^{\mathrm{foll}}\leftarrow \mathbf{0}_{n\times n}$
\For{$i=1$ to $n$}
    \For{$j=1$ to $n$}
        \If{$i=j$}
            \State \textbf{continue}
        \EndIf
        \ForAll{$(k,l)$ such that $p_i(k)=p_j(l)$}
            \If{$r_i(k)=r_j(l)$}
                \State $A^{\mathrm{foll}}_{ij}\leftarrow A^{\mathrm{foll}}_{ij}+\gamma_{\mathrm{foll}}^{k}$
            \ElsIf{$r_i(k)=0$ \textbf{or} $r_j(l)=0$}
                \State $A^{\mathrm{conf}}_{ij}\leftarrow A^{\mathrm{conf}}_{ij}+\rho_{\mathrm{conf}}\!\left(r_i(k),r_j(l)\right)\cdot\gamma_{\mathrm{conf}}^{k}$
            \ElsIf{$(r_i(k),r_j(l))~\text{are opposite}$}
                \State $A^{\mathrm{conf}}_{ij}\leftarrow A^{\mathrm{conf}}_{ij}+\gamma_{\mathrm{conf}}^{k}$
            \Else
                \State $A^{\mathrm{conf}}_{ij}\leftarrow A^{\mathrm{conf}}_{ij}+\mathbb{I}\!\left[r_i(k)=r_j(l)\right]\cdot\gamma_{\mathrm{conf}}^{k}$
            \EndIf
        \EndFor
    \EndFor
\EndFor
\State \Return $A^{\mathrm{conf}}, A^{\mathrm{foll}}$
\end{algorithmic}
\end{algorithm}

The weighted matrices $A^{\mathrm{conf}}$ and $A^{\mathrm{foll}}$ quantify pairwise path-overlap scores, but they are not directly used as dense communication graphs.
Instead, we use the similar idea as~\cite{he2025social} to select sparse and temporally stable communication partners as shown in Algorithm~\ref{alg:final_comm_graph}. 
Specifically, we first identify each agent's most relevant conflict neighbor and following neighbor:
\begin{equation}
    \begin{aligned}
        \hat{j}^{\mathrm{conf}}_i =
        \begin{cases}
        \arg\max_j A^{\mathrm{conf}}_{ij}, & \max_j A^{\mathrm{conf}}_{ij}>0,\\
        i, & \text{otherwise},
        \end{cases}\\
        \hat{j}^{\mathrm{foll}}_i =
        \begin{cases}
        \arg\max_j A^{\mathrm{foll}}_{ij}, & \max_j A^{\mathrm{foll}}_{ij}>0,\\
        i, & \text{otherwise}.
        \end{cases}
    \end{aligned}
\end{equation}
Each agent maintains a fixed conflict neighbor $f_i^{\mathrm{conf}}$ and a fixed same-direction neighbor $f_i^{\mathrm{foll}}$. 
These fixed neighbors are updated only when the previous interaction has disappeared:
\begin{equation}
    \begin{aligned}
        f_i^{g} \leftarrow
        \begin{cases}
        \hat{j}^{g}_i, & A^{g}_{i f_i^{g}}=0,\\
        f_i^{g}, & A^{g}_{i f_i^{g}}>0,
        \end{cases}
        \qquad
        g\in\{\mathrm{conf},\mathrm{foll}\}.
    \end{aligned}
\end{equation}
The final communication graphs are then constructed by adding undirected edges between each agent and its fixed neighbors:
$\mathcal{E}^{g}_{\mathrm{comm}}=\{(a_i,a_{f_i^{g}}),(a_{f_i^{g}},a_i)\mid i=1,\dots,n\}.$
This gives the final communication graphs
\begin{equation}
    \begin{aligned}
        \mathcal{G}^{\mathrm{conf}}_{\mathrm{comm}}
        =
        (\mathcal{A},\mathcal{E}^{\mathrm{conf}}_{\mathrm{comm}}),
        ~
        \mathcal{G}^{\mathrm{foll}}_{\mathrm{comm}}
        =
        (\mathcal{A},\mathcal{E}^{\mathrm{foll}}_{\mathrm{comm}}),
    \end{aligned}
\end{equation}
which are used as the communication indices in the dual-graph communication module.

Compared with directly using the dense weighted matrices $A^{\mathrm{conf}}$ and $A^{\mathrm{foll}}$, the selected communication graphs have several advantages. 
They keep communication sparse by allowing each agent to focus on its most relevant interaction, reduce noisy aggregation from weakly related agents, and improve temporal consistency through the fixed-neighbor update rule. 
As a result, the model receives a clearer and more stable communication structure while still using the weighted overlap matrices to determine which interactions are most important.

\begin{algorithm}[t]
\caption{Final Communication Graph Selection}
\label{alg:final_comm_graph}
\begin{algorithmic}[1]
\Require Weighted interaction matrices $A^{\mathrm{conf}}$, $A^{\mathrm{foll}}$; previous fixed neighbors $f^{\mathrm{conf}}$, $f^{\mathrm{foll}}$
\Ensure Final communication graphs $\mathcal{G}^{\mathrm{conf}}_{\mathrm{comm}}$ and $\mathcal{G}^{\mathrm{foll}}_{\mathrm{comm}}$

\For{$g \in \{\mathrm{conf}, \mathrm{foll}\}$}
    \For{$i = 1$ to $n$}
        \State $\hat{j}^{g}_i \gets i$
        \If{$\max_j A^{g}_{ij} > 0$}
            \State $\hat{j}^{g}_i \gets \arg\max_j A^{g}_{ij}$
        \EndIf

        \If{$A^{g}_{i f_i^{g}} = 0$}
            \State $f_i^{g} \gets \hat{j}^{g}_i$
        \EndIf
    \EndFor

    \State $\mathcal{E}^{g}_{\mathrm{comm}} \gets \emptyset$
    \For{$i = 1$ to $n$}
        \State $\mathcal{E}^{g}_{\mathrm{comm}} \gets \mathcal{E}^{g}_{\mathrm{comm}} \cup \{(a_i,a_{f_i^g}), (a_{f_i^g},a_i)\}$
    \EndFor
    \State $\mathcal{G}^{g}_{\mathrm{comm}} \gets (\mathcal{A}, \mathcal{E}^{g}_{\mathrm{comm}})$
\EndFor

\State \Return $\mathcal{G}^{\mathrm{conf}}_{\mathrm{comm}}$, $\mathcal{G}^{\mathrm{foll}}_{\mathrm{comm}}$

\end{algorithmic}
\end{algorithm}

\subsubsection{Node Feature Construction}

To expose topological constraints that may be difficult to infer directly from local observations, we represent each agent using a seven-dimensional topology-aware feature vector. 
The features are derived from the free-space topology, the agent's current and goal vertices, and an A* reference path between them. 
They characterize bottlenecks, dead-end structures, remaining travel distance, and the potential effect of an agent on the reachability and reference paths of other agents.

Let $\mathcal{G}=(\mathcal{V},\mathcal{E})$ denote the free-space graph induced by the grid map, where $\mathcal{V}$ contains the traversable cells and $\mathcal{E}$ connects four-neighboring cells. For each $v\in\mathcal{V}$, its degree is
\begin{equation}
    \begin{aligned}
        \deg_{\mathcal{G}}(v)
        =
        \left|
        \mathcal{N}_4(v)\cap\mathcal{V}
        \right|,
    \end{aligned}
\end{equation}
where $\mathcal{N}_4(v)$ denotes the four-connected grid neighborhood of $v$.
We identify a dead-end region by tracing outward from each degree-one terminal vertex through its adjacent degree-two corridor vertices until reaching a junction whose degree differs from two. The union of the terminal vertex and the traced corridor constitutes a dead-end region. We additionally identify the cut vertices of the free-space graph:
\begin{equation}
    \begin{aligned}
        \mathcal{V}_{\mathrm{cut}}
        =
        \left\{
        v\in\mathcal{V}
        \mid
        c\!\left(\mathcal{G}\setminus\{v\}\right)
        >
        c(\mathcal{G})
        \right\},
    \end{aligned}
\end{equation}
where $c(\mathcal{G})$ denotes the number of connected components in $\mathcal{G}$. Removing a cut vertex disconnects part of the free space; consequently, occupancy of such a vertex can temporarily restrict the movement and reachability of other agents. The degree, dead-end, and cut-vertex maps depend only on the static environment and can therefore be precomputed for each map.

For agent $i$, let $q_i(t)$ and $g_i$ denote its current and goal vertices, respectively, and let $P_i(t)$ denote the A* reference path from $q_i(t)$ to $g_i$. We construct the feature vector
\begin{equation}
    \begin{aligned}
        x_i(t)
        =
        \big[
        x_i^{\mathrm{cg}},
        x_i^{\mathrm{cv}},
        x_i^{\mathrm{hb}},
        x_i^{\mathrm{sb}},
        x_i^{\mathrm{dist}},
        x_i^{\mathrm{gd}},
        x_i^{\mathrm{ad}}
        \big]^\top
        \in\mathbb{R}^{7}.
    \end{aligned}
\end{equation}
For readability, the timestep index is omitted from the individual feature symbols. Table~\ref{tab:node_features} summarizes their definitions and intended roles. The agent-dependent features are updated from the current configuration and reference paths at each decision step.

\begin{table*}[t]
\centering
\caption{Topology-aware node features used by the dual-graph communication module. Here, $P_i$ denotes the A* reference path of agent $i$, and $D_i(v)$ denotes the shortest-path distance from vertex $v$ to its goal $g_i$.}
\label{tab:node_features}
\renewcommand{\arraystretch}{1.15}
\begin{tabular}{p{0.12\linewidth} p{0.18\linewidth} p{0.44\linewidth} p{0.18\linewidth}}
\toprule
\textbf{Symbol} & \textbf{Name} & \textbf{Definition} & \textbf{Role} \\
\midrule

$x_i^{\mathrm{cg}}$
& Cut-goal indicator
& Binary indicator of whether $g_i\in\mathcal{V}_{\mathrm{cut}}$.
& Identifies goals located at topological bottlenecks. \\

$x_i^{\mathrm{cv}}$
& Cut-vertex count
& Number of vertices in $\mathcal{V}_{\mathrm{cut}}$ traversed by the reference path $P_i$.
& Measures the number of bottlenecks along the reference route. \\

$x_i^{\mathrm{hb}}$
& Hard-blocking count
& If $g_i\in\mathcal{V}_{\mathrm{cut}}$, the number of other agents whose current and goal vertices become disconnected in $\mathcal{G}\setminus\{g_i\}$; otherwise zero.
& Quantifies the reachability impact of occupying a critical goal vertex. \\

$x_i^{\mathrm{sb}}$
& Soft-blocking count
& If $g_i\in\mathcal{V}_{\mathrm{cut}}$, the number of other agents $j\neq i$ whose reference paths $P_j$ contain the current vertex $q_i(t)$; otherwise zero.
& Estimates the path-overlap pressure created by the agent's current position. \\

$x_i^{\mathrm{dist}}$
& Goal distance
& Shortest-path distance $D_i(q_i(t))$, obtained from the BFS distance map rooted at $g_i$.
& Represents the agent's remaining travel distance. \\

$x_i^{\mathrm{gd}}$
& Goal-in-dead-end indicator
& Binary indicator of whether $g_i$ belongs to a dead-end region.
& Identifies goals requiring careful passage ordering or yielding. \\

$x_i^{\mathrm{ad}}$
& Agent-in-dead-end indicator
& Binary indicator of whether $q_i(t)$ belongs to a dead-end region.
& Identifies agents currently located in constrained structures. \\

\bottomrule
\end{tabular}
\end{table*}

Together, these features provide the subsequent communication module with explicit information about the structural constraints encountered by each agent and the potential effect of its movement on other agents.

\subsubsection{Dual-Graph Communication}

\begin{figure*}[t]
    \centering
    \includegraphics[width=0.95\textwidth]{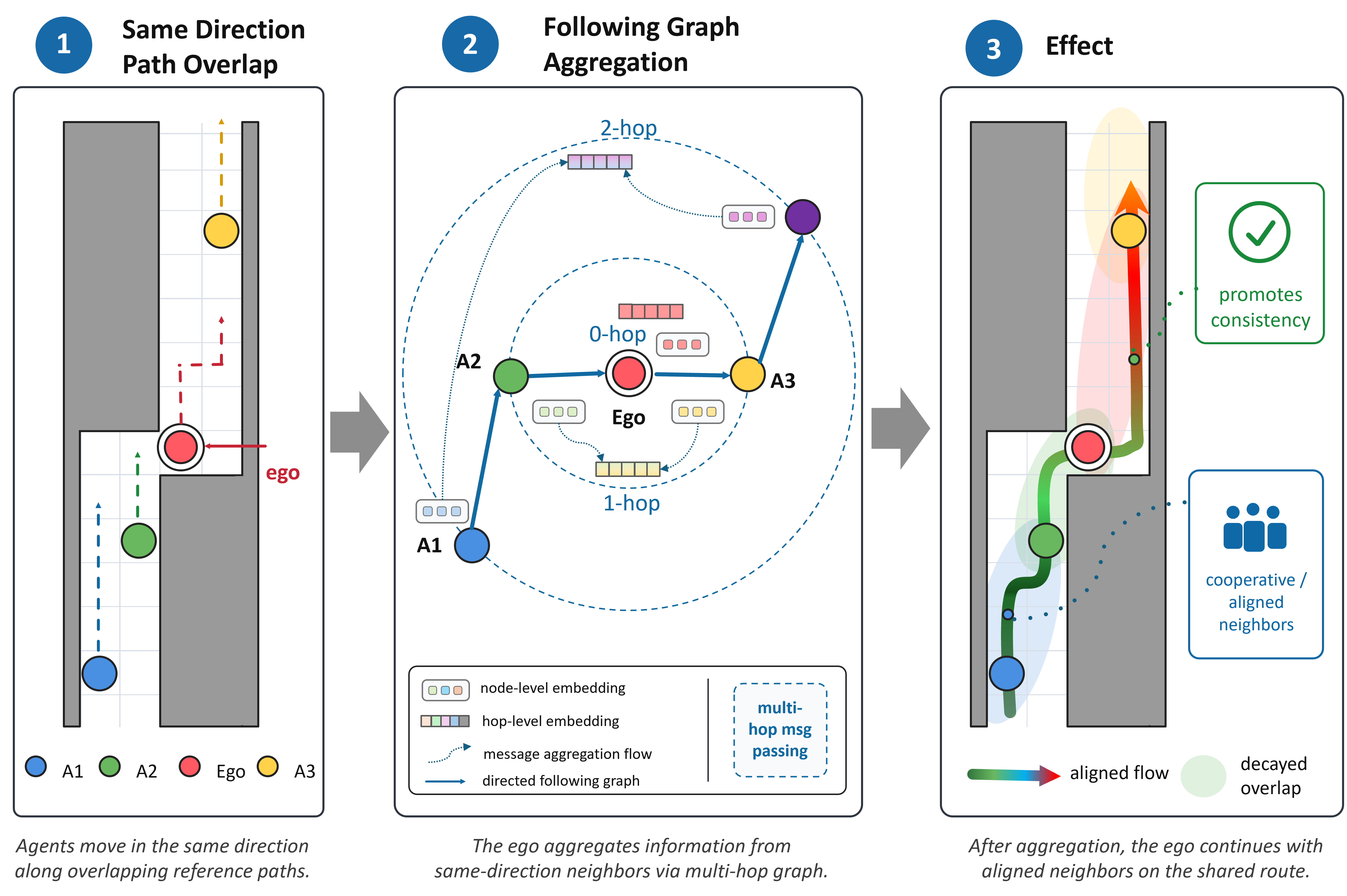}
    \caption{Following-graph communication for compatible path overlaps.
    Left: same-direction path overlaps induce directed following relations.
    Center: the ego agent aggregates representations from its one- and two-hop following neighborhoods, capturing both immediate and route-level context.
    Right: the gated multihop message supports coherent decisions along the shared route while attenuating the influence of more distant or weakly overlapping agents.}
    \label{fig:same_comm}
\end{figure*}

\begin{figure*}[t]
    \centering
    \includegraphics[width=0.95\textwidth]{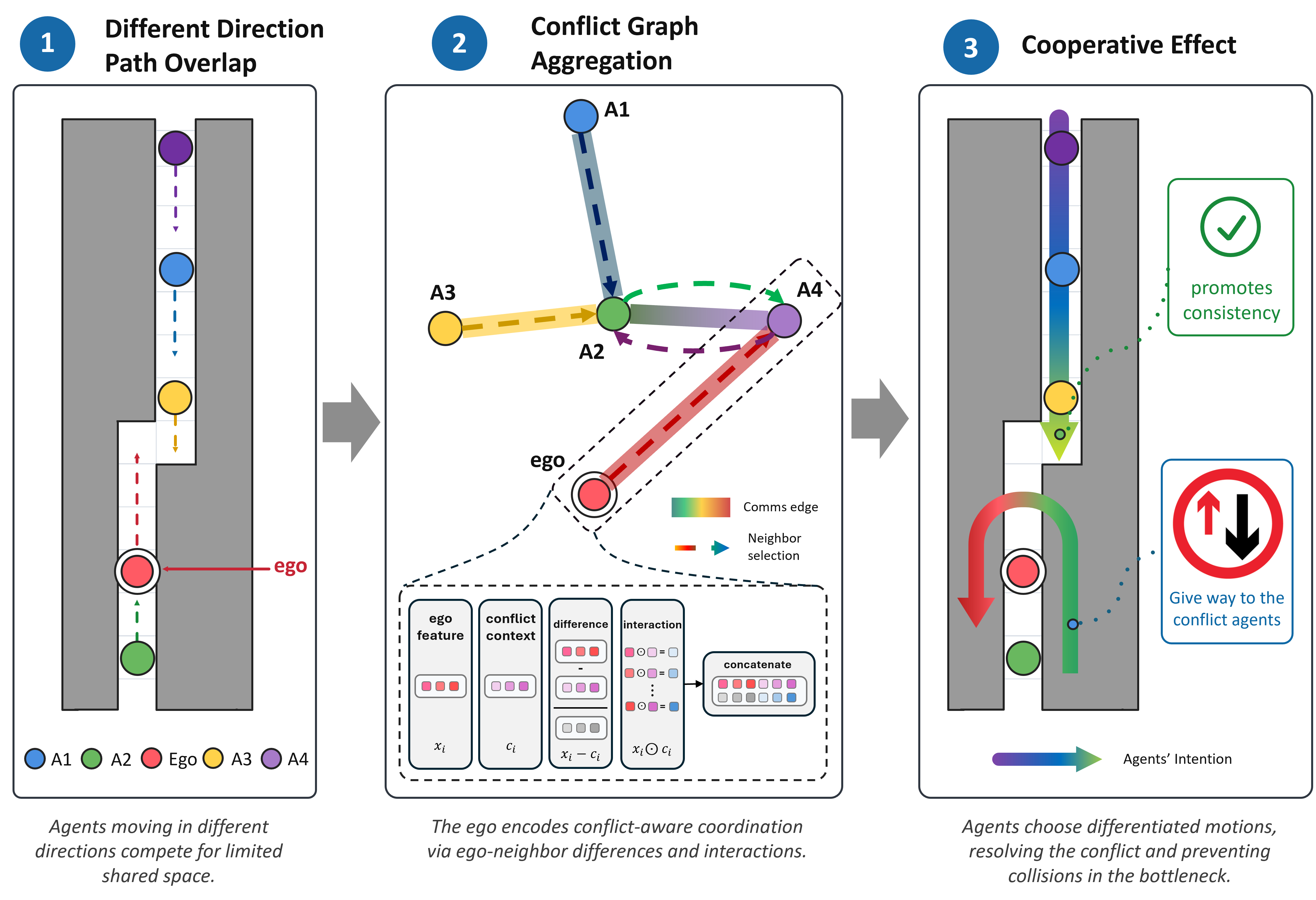}
    \caption{Conflict-graph communication for competing path overlaps.
    Left: agents approaching shared space from incompatible directions form conflict relations.
    Center: masked attention weights the selected conflict neighbors, after which the ego feature, attended context, their difference, and their element-wise interaction are encoded into a conflict message.
    Right: the resulting representation supports differentiated actions, such as proceeding or yielding, to resolve contention at the bottleneck.}
    \label{fig:conf_comm}
\end{figure*}

Given the node feature matrix $X\in\mathbb{R}^{n\times d}$ and the final communication graphs
$\mathcal{G}^{\mathrm{foll}}_{\mathrm{comm}}$ and
$\mathcal{G}^{\mathrm{conf}}_{\mathrm{comm}}$, we update the agent representations using a dual-branch graph communication module.
The underlying motivation is that path overlap alone does not determine the form of coordination required between two agents.
Agents whose reference paths are aligned or otherwise compatible benefit from sharing information that promotes coherent motion along a common route.
By contrast, agents whose reference paths compete for the same space must differentiate their decisions, for example, by determining which agent should proceed and which should yield.
We therefore process these two types of interactions using separate following and conflict branches, as illustrated in Figs.~\ref{fig:same_comm} and~\ref{fig:conf_comm}.

Let $\tilde{A}^{\mathrm{foll}}$ and $\tilde{A}^{\mathrm{conf}}$ denote the binary adjacency matrices induced by
$\mathcal{G}^{\mathrm{foll}}_{\mathrm{comm}}$ and
$\mathcal{G}^{\mathrm{conf}}_{\mathrm{comm}}$, respectively.
We adopt the convention that $\tilde{A}_{ij}=1$ permits agent $i$ to aggregate information from agent $j$.

\paragraph{Following-graph aggregation.}
The following branch performs cooperative message aggregation among agents with compatible path structures.
For these agents, information should propagate smoothly beyond immediate neighbors so that agents moving along the same route can share route-level context.
We therefore adopt a multihop aggregation scheme~\cite{he2025social} inspired by Hop2Token~\cite{chen2022nagphormer}.

We first add self-loops and construct a degree-normalized adjacency matrix:
\begin{equation}
    \begin{aligned}
        \hat{A}^{\mathrm{foll}}
        &=
        \tilde{A}^{\mathrm{foll}}+I, \\
        D^{\mathrm{foll}}_{ii}
        &=
        \sum_j \hat{A}^{\mathrm{foll}}_{ij}, \\
        \bar{A}^{\mathrm{foll}}
        &=
        \left(D^{\mathrm{foll}}\right)^{-\frac{1}{2}}
        \hat{A}^{\mathrm{foll}}
        \left(D^{\mathrm{foll}}\right)^{-\frac{1}{2}}.
    \end{aligned}
\end{equation}
The first- and second-order messages are then computed as
\begin{equation}
    \begin{aligned}
        M^{\mathrm{foll}}_1
        &=
        \bar{A}^{\mathrm{foll}}\sigma(XW_1),\\
        M^{\mathrm{foll}}_2
        &=
        \bar{A}^{\mathrm{foll}}M^{\mathrm{foll}}_1,
    \end{aligned}
\end{equation}
where $W_1$ is a learnable projection matrix and $\sigma(\cdot)$ denotes a nonlinear activation function.
Because the normalized adjacency matrix contains self-loops, $M^{\mathrm{foll}}_1$ and $M^{\mathrm{foll}}_2$ encode information from neighborhoods of up to one and two hops, respectively.
The former captures interactions with immediate following neighbors, whereas the latter propagates information along short chains of path-compatible agents.

A feature-wise gate adaptively combines the two message scales:
\begin{equation}
    \begin{aligned}
        G^{\mathrm{foll}}
        &=
        \operatorname{sigmoid}
        \left(
        f_{\mathrm{gate}}
        \left(
        [X\Vert M^{\mathrm{foll}}_1
        \Vert M^{\mathrm{foll}}_2]
        \right)
        \right),\\
        M^{\mathrm{foll}}
        &=
        G^{\mathrm{foll}}\odot M^{\mathrm{foll}}_1
        +
        \left(\mathbf{1}-G^{\mathrm{foll}}\right)
        \odot f_{\mathrm{foll}}
        \left(M^{\mathrm{foll}}_2\right),
    \end{aligned}
\end{equation}
where $\Vert$ denotes feature concatenation, $\odot$ denotes element-wise multiplication, and $f_{\mathrm{gate}}$ and $f_{\mathrm{foll}}$ are learnable nonlinear transformations.
The following-aware representation is obtained using a residual connection followed by layer normalization:
\begin{equation}
    \begin{aligned}
        H^{\mathrm{foll}}
        =
        \operatorname{LN}
        \left(
        X+M^{\mathrm{foll}}
        \right).
    \end{aligned}
\end{equation}
This branch provides each agent with multiscale context from path-compatible agents, supporting coherent decisions along shared routes.

\paragraph{Conflict-graph aggregation.}
The conflict branch models interactions among agents whose reference paths compete for shared space.
Unlike the following branch, it should not smooth the representations of neighboring agents indiscriminately.
Instead, it must preserve the relative information needed to support asymmetric decisions, such as proceeding, waiting, or yielding.
We therefore apply masked attention over the conflict graph, restricting each agent to its selected conflict neighbors.

The query, key, and value matrices are computed as
\begin{equation}
    \begin{aligned}
        Q=XW_Q,\qquad
        K=XW_K,\qquad
        V=XW_V,
    \end{aligned}
\end{equation}
where $Q,K\in\mathbb{R}^{n\times d_k}$ and
$V\in\mathbb{R}^{n\times d}$.
For agent $i$, let
\begin{equation}
    \mathcal{N}^{\mathrm{conf}}_i
    =
    \left\{
    j \mid \tilde{A}^{\mathrm{conf}}_{ij}>0
    \right\}
\end{equation}
denote its selected conflict neighbors.
The masked attention scores and normalized weights are
\begin{equation}
    \begin{aligned}
        s_{ij}
        &=
        \frac{q_i^\top k_j}{\sqrt{d_k}},
        \qquad j\in\mathcal{N}^{\mathrm{conf}}_i,\\
        \alpha_{ij}
        &=
        \frac{\exp(s_{ij})}
        {\sum_{\ell\in\mathcal{N}^{\mathrm{conf}}_i}
        \exp(s_{i\ell})}.
    \end{aligned}
\end{equation}
The resulting conflict context is
\begin{equation}
    \begin{aligned}
        c_i
        =
        \sum_{j\in\mathcal{N}^{\mathrm{conf}}_i}
        \alpha_{ij}v_j.
    \end{aligned}
\end{equation}
If no external conflict neighbor is selected, we set
$\mathcal{N}^{\mathrm{conf}}_i=\{i\}$.
This self-edge avoids an empty attention operation and reduces the context to the agent's own projected representation.

To retain information that distinguishes the ego agent from its competitors, the conflict message is not constructed by directly averaging their features.
Instead, we explicitly encode the ego feature, the attended conflict context, their difference, and their element-wise interaction:
\begin{equation}
    \begin{aligned}
        m_i^{\mathrm{conf}}
        =
        f_{\mathrm{conf}}
        \left(
        [x_i\Vert c_i\Vert
        (x_i-c_i)\Vert
        (x_i\odot c_i)]
        \right).
    \end{aligned}
\end{equation}
The difference term exposes relative feature discrepancies, whereas the interaction term captures feature-wise compatibility between the ego agent and its conflict context.
The conflict-aware representation is then computed as
\begin{equation}
    \begin{aligned}
        h_i^{\mathrm{conf}}
        =
        \operatorname{LN}
        \left(
        x_i+m_i^{\mathrm{conf}}
        \right).
    \end{aligned}
\end{equation}
Rather than directly enforcing a particular yielding rule, this representation provides the downstream policy with the relational information required to differentiate the actions of competing agents.

\paragraph{Adaptive fusion.}
Finally, the two branches are combined using independent feature-wise gates.
Let $m_i^{\mathrm{foll}}$ and $h_i^{\mathrm{foll}}$ denote the $i$th rows of
$M^{\mathrm{foll}}$ and $H^{\mathrm{foll}}$, respectively.
The fusion gates are computed as
\begin{equation}
    \begin{aligned}
        [g_i^{\mathrm{foll}}\Vert g_i^{\mathrm{conf}}]
        =
        \operatorname{sigmoid}
        \left(
        f_{\mathrm{fuse}}
        \left(
        [x_i\Vert h_i^{\mathrm{foll}}
        \Vert h_i^{\mathrm{conf}}]
        \right)
        \right),
    \end{aligned}
\end{equation}
where
$g_i^{\mathrm{foll}},g_i^{\mathrm{conf}}\in(0,1)^d$
independently modulate the following and conflict messages.
The final representation is
\begin{equation}
    \begin{aligned}
        \tilde{x}_i
        =
        \operatorname{LN}
        \left(
        x_i
        +
        g_i^{\mathrm{foll}}\odot m_i^{\mathrm{foll}}
        +
        g_i^{\mathrm{conf}}\odot m_i^{\mathrm{conf}}
        \right).
    \end{aligned}
\end{equation}
The updated representation $\tilde{x}_i$ is subsequently passed to the policy network for action prediction.

Overall, the dual-graph module applies distinct communication mechanisms to two qualitatively different interaction structures.
The following branch propagates multihop context among agents with compatible reference paths, whereas the conflict branch preserves ego-neighbor differences that are important for priority-sensitive coordination.
The adaptive fusion gates allow the policy to modulate both information sources for each agent at every decision step.

\subsection{LaCAM3-Guided Confidence Boosting}

Although reinforcement learning allows agents to acquire decentralized coordination strategies through interaction, the learned policy may remain uncertain in states that require precise joint decisions. 
At bottlenecks, dead ends, and congested intersections, for example, sampling from a high-entropy action distribution can produce inconsistent decisions, leading to unnecessary waiting, blocking, or repeated conflicts. 
We therefore use LaCAM3 as a training-only expert within a confidence-triggered intervention framework. 
LaCAM3 provides coordinated action guidance for uncertain agents and expert supervision that encourages the learned policy to assign greater probability to decisive actions.

\paragraph{Entropy-based uncertainty detection}
Let
$\pi_\theta^i(\cdot\mid o_t^i)$
denote the action distribution of agent $i$ at timestep $t$, conditioned on its observation $o_t^i$. We quantify the uncertainty of the policy using its action entropy:
\begin{equation}
    \begin{aligned}
        \mathcal{H}_t^i
        =
        -\sum_{u\in\mathcal{U}}
        \pi_\theta^i(u\mid o_t^i)
        \log\left(
        \pi_\theta^i(u\mid o_t^i)
        +\varepsilon_{\mathrm{num}}
        \right),
    \end{aligned}
\end{equation}
where $\mathcal{U}$ is the discrete action space and
$\varepsilon_{\mathrm{num}}$ is a small constant for numerical stability. A high entropy indicates that the policy has not developed a clear preference among the candidate actions. The set of uncertain agents is therefore defined as
\begin{equation}
    \begin{aligned}
        \mathcal{I}_t^{\mathrm{unc}}
        =
        \left\{
        i\mid \mathcal{H}_t^i>\eta
        \right\},
    \end{aligned}
\end{equation}
where $\eta$ is a fixed uncertainty threshold.

For the five-action policy considered in this work, we set $\eta$ to the entropy of the reference distribution
\begin{equation}
    \begin{aligned}
        p_{\mathrm{ref}}
        =
        [0.9,0.025,0.025,0.025,0.025].
    \end{aligned}
\end{equation}
This distribution represents a confident policy with one dominant action and a small residual probability assigned to each alternative. Using natural logarithms, the resulting threshold is approximately $\eta=0.464$. An agent is identified as uncertain whenever its action entropy exceeds this reference value.

\paragraph{Confidence-triggered expert intervention}
Given the obstacle map, current agent positions, and goal locations, LaCAM3 is queried to generate a collision-free joint plan. Let $u_t^{*,i}$ denote the expert action assigned to agent $i$ at timestep $t$. The expert action is converted into a label-smoothed categorical target:
\begin{equation}
    \begin{aligned}
        \pi_t^{*,i}(u)
        =
        \begin{cases}
            1-(|\mathcal{U}|-1)\varepsilon_{\mathrm{E}},
            & u=u_t^{*,i},\\
            \varepsilon_{\mathrm{E}},
            & u\neq u_t^{*,i},
        \end{cases}
    \end{aligned}
\end{equation}
where $\varepsilon_{\mathrm{E}}=0.025$. Because
$|\mathcal{U}|=5$, the target assigns a probability of $0.9$ to the expert action and $0.025$ to each alternative. This label smoothing preserves a clear preference for the expert action without imposing a hard one-hot target.

Training consists of two phases. During the first phase, the policy is trained exclusively using reinforcement learning, without expert intervention or imitation loss. This warm-up phase allows the agents to first acquire basic navigation and coordination behaviors through environmental interaction. During the second phase, LaCAM3 guidance and expert supervision are introduced to refine uncertain decisions.

Within the second training phase, each episode is divided into an online-replanning segment and a cached-plan segment. During the online-replanning segment, LaCAM3 is invoked at each timestep from the current joint state. The expert action is executed only for agents identified as uncertain:
\begin{equation}
    \begin{aligned}
        \tilde{u}_t^i
        =
        \begin{cases}
            u_t^{*,i},
            & i\in\mathcal{I}_t^{\mathrm{unc}},\\
            u_t^i,
            & i\notin\mathcal{I}_t^{\mathrm{unc}},
        \end{cases}
    \end{aligned}
\end{equation}
where $u_t^i$ is sampled from the learned policy and
$\tilde{u}_t^i$ is the action executed in the environment. This selective intervention retains confident decisions acquired through reinforcement learning while replacing uncertain decisions with coordinated expert actions.

The agents that remain unresolved later in an episode typically face more difficult residual conflicts. At a predefined switching timestep $t_{\mathrm{sw}}$, LaCAM3 is therefore invoked once more from the current joint state, and its joint plan is cached for the remainder of the episode. Let
$\mathcal{I}_{t_{\mathrm{sw}}}^{\mathrm{rem}}$
denote the agents that have not reached their goals at
$t_{\mathrm{sw}}$, and let
$\bar{u}_t^{*,i}$ denote the action obtained from the cached expert plan. During this segment, the executed actions are
\begin{equation}
    \begin{aligned}
        \tilde{u}_t^i
        =
        \begin{cases}
            \bar{u}_t^{*,i},
            & i\in\mathcal{I}_{t_{\mathrm{sw}}}^{\mathrm{rem}},\\
            u_t^i,
            & \text{otherwise},
        \end{cases}
        \qquad t\geq t_{\mathrm{sw}}.
    \end{aligned}
\end{equation}
This stronger intervention exposes the policy to coordinated completion behavior in difficult late-episode states while avoiding repeated replanning during the remaining timesteps.

\paragraph{Expert-supervised policy refinement}
In addition to action intervention, we introduce a behavior cloning loss during the second training phase. Let
$\mathcal{D}_{\mathrm{E}}$
denote the set of agent-timestep pairs for which expert supervision is enabled. This set includes entropy-triggered interventions during the online-replanning segment and unresolved agents guided by the cached plan during the late-episode segment.

Because the action space is categorical, we minimize the cross-entropy between the smoothed expert target and the learned action distribution:
\begin{equation}
    \begin{aligned}
        \mathcal{L}_{\mathrm{BC}}
        =
        -\mathbb{E}_{(t,i)\sim\mathcal{D}_{\mathrm{E}}}
        \left[
        \sum_{u\in\mathcal{U}}
        \pi_t^{*,i}(u)
        \log\left(
        \pi_\theta^i(u\mid o_t^i)
        +\varepsilon_{\mathrm{num}}
        \right)
        \right].
    \end{aligned}
\end{equation}
This loss increases the probability assigned to the expert action while reducing the relative probability of competing actions. Compared with a hard one-hot target, the smoothed distribution discourages excessive overconfidence and provides a less abrupt supervisory signal.

The overall training loss during the second phase is
\begin{equation}
    \begin{aligned}
        \mathcal{L}
        =
        \mathcal{L}_{\mathrm{RL}}
        +
        \lambda_{\mathrm{BC}}
        \mathcal{L}_{\mathrm{BC}},
    \end{aligned}
\end{equation}
where $\lambda_{\mathrm{BC}}$ controls the strength of expert supervision. During the first phase,
$\lambda_{\mathrm{BC}}=0$, and no LaCAM3 labels are used.

Unlike conventional behavior cloning from a fixed offline dataset, the proposed approach queries LaCAM3 at states encountered by the evolving policy. The planner therefore serves two complementary roles: it intervenes when the learned action distribution is uncertain and provides online expert targets for policy refinement. Importantly, LaCAM3 is used only during training; at inference time, action selection is performed entirely by the learned decentralized policy. The resulting policy is encouraged to make more decisive and coordinated decisions in challenging MAPF configurations, particularly near bottlenecks and during late-episode residual conflicts.

\subsection{Priority-Aware PIBT-based Action Refinement}
\label{subsec:pibt_refinement}

Several learning-based MAPF methods employ PIBT as an action shield~\cite{veerapaneni2024improving,jiang2025deploying,jain2026pairwise}. 
In this setting, a neural policy proposes an action for each agent, after which PIBT constructs an executable joint action by resolving vertex and edge-swap conflicts and excluding moves into obstacles. 
Although this approach improves execution safety, it is primarily reactive to conflicts in the proposed joint action. 
When an agent's preferred action cannot be executed, the ordering of its fallback actions may be only weakly informed by the learned policy and can consequently depend heavily on heuristic or random tie-breaking.

We retain the collision-resolution mechanism of PIBT but extend it into a priority-aware action refinement module. 
Specifically, we modify both the initialization of agent priorities and the ordering of candidate actions. 
These modifications incorporate persistent temporal information, learned coordination priorities, and goal-directed action preferences into the shielding process. 
This is particularly important in dense MAPF instances, where repeated local conflicts and arbitrary fallback decisions can lead to oscillatory behavior or prolonged blocking.

\paragraph{Persistent agent priority initialization}
Because PIBT is invoked independently at every timestep as a one-step refinement module, priority information maintained internally during one invocation does not naturally persist to the next. 
We therefore maintain an environment-side age variable $\tau_i(t)$ for each agent $i$. It is initialized as $\tau_i(0)=0$ and updated according to
\begin{equation}
    \begin{aligned}
        \tau_i(t+1)
        =
        \begin{cases}
            0,
            & q_i(t+1)=g_i,\\
            \tau_i(t)+1,
            & q_i(t+1)\neq g_i,
        \end{cases}
    \end{aligned}
\end{equation}
where $q_i(t)$ and $g_i$ denote the current and goal vertices of agent $i$, respectively. 
The age variable therefore records the number of consecutive timesteps for which the agent has remained unfinished and preserves this information across successive PIBT calls.

At each timestep, the initial PIBT priority of agent $i$ is computed as
\begin{equation}
    \begin{aligned}
        \rho_i(t)
        =
        w_{\mathrm{age}}\tau_i(t)
        +
        w_{\mathrm{prio}}s_{\theta,i}^{\mathrm{prio}}(o_t^i)
        +
        w_{\mathrm{dist}}
        \frac{D_i(q_i(t))}{|\mathcal{V}|},
    \end{aligned}
\end{equation}
where $s_{\theta,i}^{\mathrm{prio}}(o_t^i)$ is the scalar priority predicted by the neural network~\cite{he2025social}, $D_i(v)$ is the shortest-path distance from vertex $v$ to goal $g_i$ on the static map, and $|\mathcal{V}|$ is the number of traversable vertices. The coefficients $w_{\mathrm{age}}$, $w_{\mathrm{prio}}$, and $w_{\mathrm{dist}}$ control the contributions of the three terms. Agents are processed in descending order of $\rho_i(t)$.

The persistent age term carries temporal information across PIBT calls, while the learned priority communicates the policy's assessment of the current coordination state. The normalized distance term provides an additional state-dependent criterion for agents with similar age and learned priority. With $w_{\mathrm{dist}}>0$, this term gives precedence to agents with longer remaining paths; if agents closer to their goals are intended to receive precedence, the sign of this term should instead be reversed.

\paragraph{Policy-aware action candidate ordering}
Rather than providing PIBT with only a discrete action, we use an action-preference vector that preserves the policy's relative preferences over fallback actions. 
Let $p_i(u)$ denote the probability assigned to action $u$ by agent $i$, and let $\hat{u}_i$ denote the discrete action selected by the policy. 
We first clip the probabilities to nonnegative values and then assign the selected action a score strictly greater than all other actions:
\begin{equation}
    \begin{aligned}
        \bar{p}_i(u)
        =
        \begin{cases}
            \displaystyle
            \max_{u'\in\mathcal{U}}
            [p_i(u')]_+ + 1,
            & u=\hat{u}_i,\\[1mm]
            [p_i(u)]_+,
            & u\neq\hat{u}_i,
        \end{cases}
    \end{aligned}
\end{equation}
where $[x]_+=\max(x,0)$. The resulting scores are normalized before being passed to PIBT:
\begin{equation}
    \begin{aligned}
        \tilde{p}_i(u)
        =
        \frac{\bar{p}_i(u)}
        {\sum_{u'\in\mathcal{U}}\bar{p}_i(u')}.
    \end{aligned}
\end{equation}
Because normalization preserves the ordering of the scores, $\hat{u}_i$ is always the most preferred action.
If it cannot be assigned safely, PIBT selects among the remaining actions according to their probabilities under the learned policy while applying its standard priority-inheritance and collision-resolution procedure. 
This design preserves the policy's primary decision while retaining its full action distribution to guide fallback selection, thereby reducing reliance on arbitrary tie-breaking.

\section{Results}

In this section, we present a comprehensive evaluation of PRIMAL3. 
We first compare our method with representative traditional and learning-based MAPF baselines under different map layouts and agent densities. 
We then test PRIMAL3 in ultra-large-scale teams to verify its scalability.
We finally perform ablation studies to quantify the effect of each major component of our framework, including the following graph, conflict graph, and our proposed LaCAM3 guidance.

\subsection{Training Setup}

All experiments are run on a workstation with an AMD Ryzen 9 9950X3D 16-Core CPU, 64 GB RAM, and an NVIDIA RTX 5090 GPU with 32 GB memory. 
During training, we adopt a curriculum over the number of agents. 
The policy is first trained on 8-agent instances and is then exposed to larger teams, where the number of agents is randomly sampled from $\{10,12,14\}$. 
For each episode, the map size is uniformly sampled from $[10,30]$, and both the initial positions and goal locations of all agents are randomly generated on valid free cells. 
An episode ends once all agents reach their goals or when the horizon exceeds 256 timesteps. 
The model is optimized using Adam with a learning rate of $1\times10^{-5}$. 
During execution stage, we consider both random maps and maze maps. 
For random maps, the obstacle density is set to approximately $0.3$, while for maze maps it is set to approximately $0.5$. 
And the episode horizon is extended to 512 timesteps.

\subsection{Comparison Results}

\begin{figure*}[t]
    \centering
    \includegraphics[width=\textwidth]{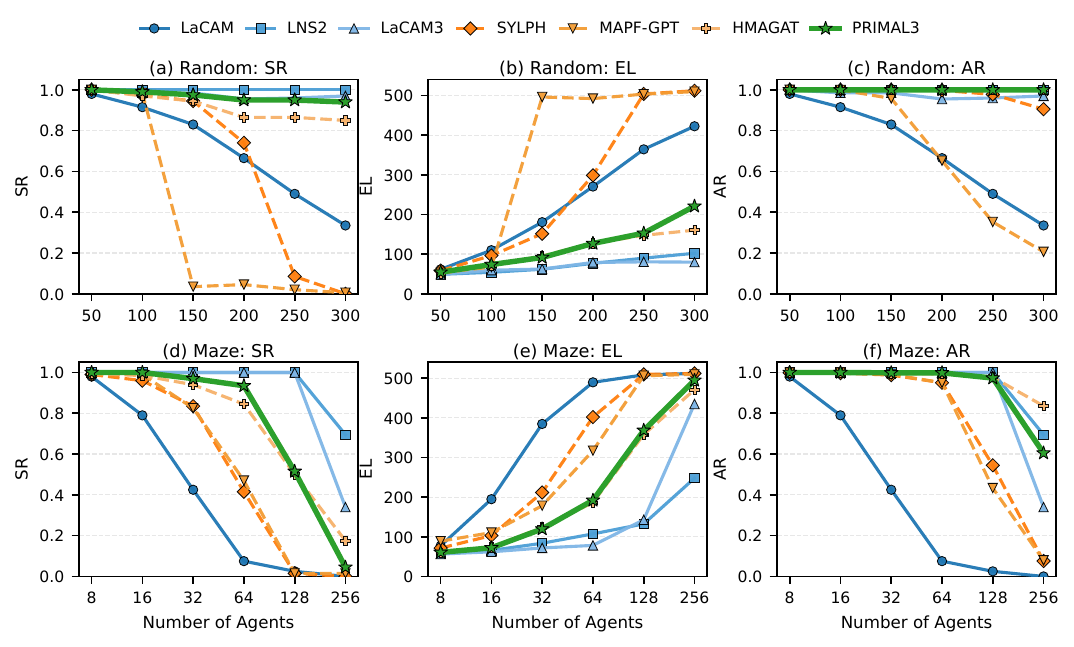}
    \caption{Comparison results on $32\times32$ random and maze maps across varying numbers of agents. We report success rate (SR), episode length (EL), and arrival rate (AR). Higher SR and AR are better, whereas lower EL is preferred. PRIMAL3 is highlighted in green.}
    \label{fig:comparison_results}
\end{figure*}

We compare PRIMAL3 with representative search-based MAPF solvers, including LaCAM~\cite{okumura2023lacam}, LaCAM3~\cite{okumura2023engineering}, and LNS2~\cite{li2022mapf}, as well as recent learning-based methods, including MAPF-GPT~\cite{andreychuk2025mapf}, SYLPH~\cite{he2025social}, and HMAGAT~\cite{jain2026pairwise}. 
We conduct evaluations on two classes of $32\times32$ maps.
The first class consists of random maps with obstacle densities of approximately $20$--$30\%$, on which we evaluate $50$, $100$, $150$, $200$, $250$, and $300$ agents. 
The second class consists of maze maps with obstacle densities of approximately $40$--$50\%$, on which we evaluate $8$, $16$, $32$, $64$, $128$, and $256$ agents. 
Each configuration contains $200$ test instances. 
Aligned with prior works, search-based solvers are given a planning-time limit of $30$ seconds per instance, while learning-based methods are evaluated with a maximum execution horizon of $512$ timesteps. 
We report success rate (SR), episode length (EL), and arrival rate (AR), where higher SR and AR are preferred and lower EL is better. 
The complete results are presented in Fig~\ref{fig:comparison_results}.

On random maps, PRIMAL3 achieves the strongest overall performance among the learning-based methods.
Excitingly, its performance is almost on par with state-of-the-art traditional search-based methods.
Its SR remains high as the number of agents increases, reaching $1.00$, $0.99$, $0.985$, $0.95$, $0.95$, and $0.92$ for $50$ to $300$ agents, respectively. 
PRIMAL3 consistently outperforms HMAGAT, the strongest learning-based baseline in most random-map settings, with the advantage becoming particularly evident in larger teams.
SYLPH remains competitive at small and moderate scales but degrades sharply as congestion increases, while the performance of MAPF-GPT drops substantially beyond $100$ agents. 
In addition to its high SR, PRIMAL3 maintains an AR close to $1.0$ across all random-map settings. Thus, even when an instance is not completely solved within the execution horizon, nearly all agents still reach their goals.

PRIMAL3 does not always yield the lowest EL among the learning-based methods. In particular, HMAGAT achieves lower EL at $250$ and $300$ agents, but also solves fewer instances and produces lower AR. EL should therefore be interpreted jointly with SR and AR: PRIMAL3 successfully resolves a larger fraction of instances, although some require additional coordination steps. This behavior is consistent with a policy that employs waiting or detouring actions to resolve congestion rather than repeatedly following locally shortest paths that can lead to blocking or oscillatory behavior in crowded environments. Overall, the combination of high SR, near-perfect AR, and moderate EL indicates that PRIMAL3 improves solvability and coordination robustness at the cost of longer execution in some cases.

Maze maps are substantially more challenging because their narrow passages and stronger topological constraints require more coordinated interactions among agents. 
PRIMAL3 achieves the best or near-best performance among the learning-based methods in the small- and medium-scale regimes. 
Specifically, its SR reaches $1.00$, $1.00$, $0.975$, and $0.935$ for $8$, $16$, $32$, and $64$ agents, respectively, outperforming SYLPH, MAPF-GPT, and HMAGAT at these scales. 
Search-based solvers such as LNS2 and LaCAM3 also maintain near-perfect SR in this regime, indicating that these instances remain broadly solvable despite their increased difficulty. 
Within this practically meaningful comparison range, PRIMAL3 demonstrates consistently stronger learned coordination than the other learning-based methods.

At $128$ agents, the SR of PRIMAL3 decreases to approximately $0.5$, whereas its AR remains close to $1.0$. 
This indicates that, although PRIMAL3 cannot always bring every agent to its goal within the horizon, it still successfully routes the vast majority of agents. 
At the most congested setting of $256$ agents, HMAGAT achieves higher SR and AR than PRIMAL3. 
Nevertheless, all learning-based methods exhibit low SR in this extreme regime, highlighting a shared limitation of current reactive MAPF policies under severe maze congestion. 
Overall, PRIMAL3 performs most favorably in regimes where learned policies can still achieve reliable instance-level completion, while extreme congestion remains an open challenge.

Search-based solvers remain stronger under the current evaluation protocol, benefiting from centralized planning and up to $30$ seconds of computation per instance. 
On random maps, LNS2 achieves perfect SR across all tested agent populations, while LaCAM3 also maintains high SR and low EL. 
On maze maps, LNS2 and LaCAM3 remain effective at larger agent populations than the learning-based policies. 
Nevertheless, among the learning-based methods, PRIMAL3 delivers the strongest and most consistent performance across the evaluated random and maze maps, substantially narrowing the gap between reactive learned policies and centralized MAPF solvers. 
These results support the overall effectiveness of combining topology-aware node features, dual-graph communication, and LaCAM3-guided confidence boosting in PRIMAL3.

\subsection{Ultra-Large-Scale Experiments}
\label{sec:ultra_large_scale}

To evaluate the scalability of PRIMAL3, we increase the team size while approximately maintaining an obstacle density of $0.2$ and an agent density of $0.2$ over all grid cells ($0.25$ over free cells).
We evaluate
$N\in\{1000,2000,3000,4000,5000,7500,10{,}000\}$
agents on square maps with corresponding side lengths
$\{72,101,124,143,160,196,226\}$.
The maximum episode horizons are $1024$, $1536$, and $2048$ steps for 1000--3000, 4000--5000, and 7500--$10{,}000$ agents, respectively.
We focus on HMAGAT because it is the strongest learning-based baseline in our preceding experiments and is explicitly designed to capture group interactions through hypergraph-based communication~\cite{jain2026pairwise}.

\begin{figure*}[t]
\centering
\includegraphics[width=0.88\textwidth]{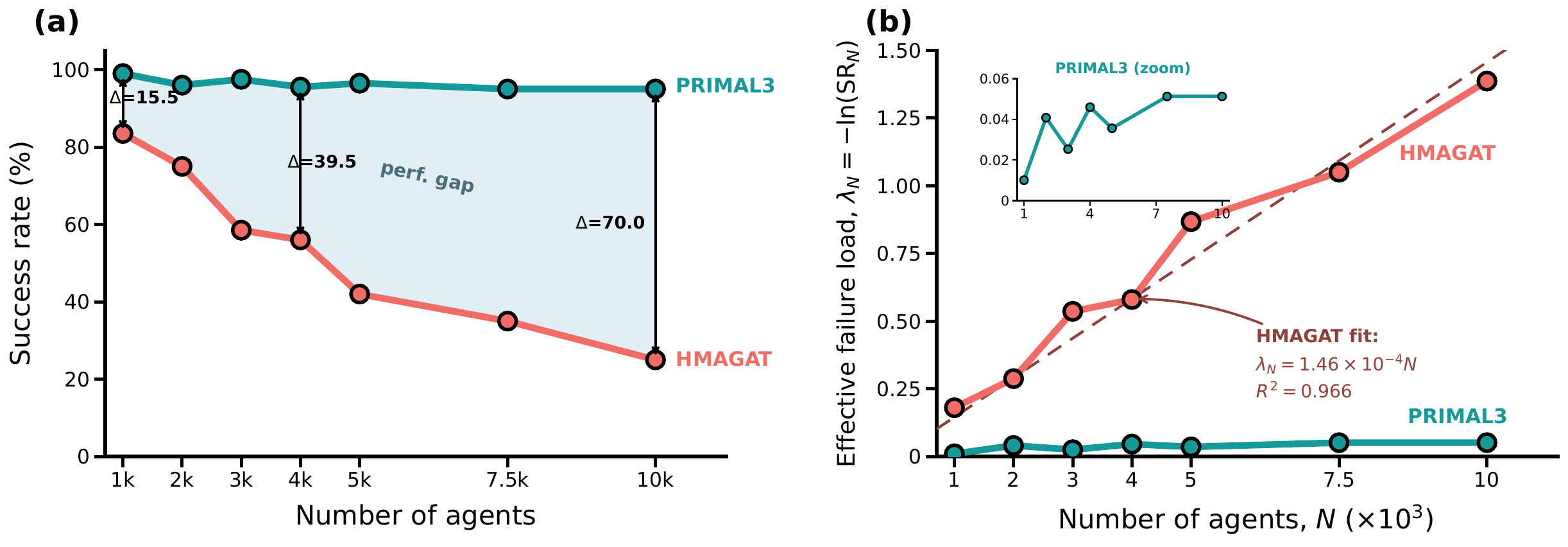}
\caption{
Ultra-large-scale evaluation of PRIMAL3 and HMAGAT.
(a) Instance-level success rate, where success requires all agents to reach their goals.
(b) Effective failure load $\lambda_N=-\ln\mathrm{SR}_N$.
The dashed line shows a least-squares fit for HMAGAT constrained through the origin,
$\lambda_N=1.46\times10^{-4}N$ (centered $R^2=0.966$).
}
\label{fig:ultra_large_scale}
\end{figure*}

As shown in Fig~\ref{fig:ultra_large_scale}(a), PRIMAL3 maintains success rates between $95.0\%$ and $99.0\%$ across all evaluated scales, achieving $95.0\%$ even with $10{,}000$ agents.
In contrast, HMAGAT's success rate decreases from $83.5\%$ at 1000 agents to $25.0\%$ at $10{,}000$ agents.
Although the local obstacle and agent densities remain approximately fixed, larger maps involve longer paths and execution horizons, as well as a larger total number of coordination events.
Moreover, because an instance succeeds only when every agent reaches its goal, even a stable residual failure risk can accumulate rapidly with the team size.

We therefore examine the effective failure load
$\lambda_N=-\ln\mathrm{SR}_N$.\footnote{
The transformation $H=-\ln S$ is the standard cumulative-hazard representation in survival and reliability analysis~\cite{nelson1972theory,kalbfleisch2002statistical}.
Here, $\lambda_N$ is defined as a descriptive scaling statistic and does not require any assumption about dependence between failures.
Under a simple reference model in which agents contribute independent and scale-invariant failure risks $\epsilon$, $\mathrm{SR}_N=(1-\epsilon)^N$ and $\lambda_N=-N\ln(1-\epsilon)\approx N\epsilon$.
Therefore, linear growth of $\lambda_N$ with $N$ is consistent with a
constant effective per-agent failure hazard, although it does not establish that the actual MAPF failures are independent.
}
As shown in Fig~\ref{fig:ultra_large_scale}(b), HMAGAT's $\lambda_N$ grows approximately linearly with $N$, indicating that its declining instance-level success rate is consistent with the accumulation of a nearly constant effective per-agent failure hazard.
In contrast, PRIMAL3 keeps $\lambda_N$ below $0.052$, and its effective per-agent hazard, $\lambda_N/N$, is approximately 16 times lower on average across the evaluated scales.
The widening success-rate gap therefore reflects the all-agent success criterion amplifying a genuine difference in residual coordination reliability, rather than a sudden deterioration of HMAGAT's local policy at larger scales.

The lower failure hazard of PRIMAL3 is consistent with its task-conditioned and temporally persistent coordination mechanisms.
HMAGAT constructs spatial hyperedges using a fixed communication radius and message-passing depth, whereas PRIMAL3 explicitly separates path-conditioned conflict and following relations.
Because each agent selects at most one partner in each graph, each symmetrized graph contains at most $N$ edges and consequently has an average degree of at most $2$, independent of the team size.
More importantly, PRIMAL3 allocates this bounded communication budget to path-relevant interactions, and the selected conflict partner is retained until the corresponding path overlap is resolved.
Together with topology-aware observations, uncertainty-triggered LaCAM3 guidance, and priority-aware PIBT refinement, these mechanisms reduce persistent blocking and temporally inconsistent conflict resolution over long executions.
This interpretation is consistent with HMAGAT's reported failure analysis, in which livelock and exhaustion of the maximum timestep constitute the dominant failure modes~\cite{jain2026pairwise}.

Finally, PRIMAL3 successfully completes two additional single-instance stress tests (with the same obstacle density and agent density) with $50{,}000$ and $100{,}000$ agents, demonstrating the feasibility of coordination at a scale approaching city-level multi-agent systems.

\subsection{Ablation Study}

\subsubsection{Components Ablation}
\label{sec:components_ablation}

\begin{figure*}[t]
    \centering
    \includegraphics[width=\linewidth]{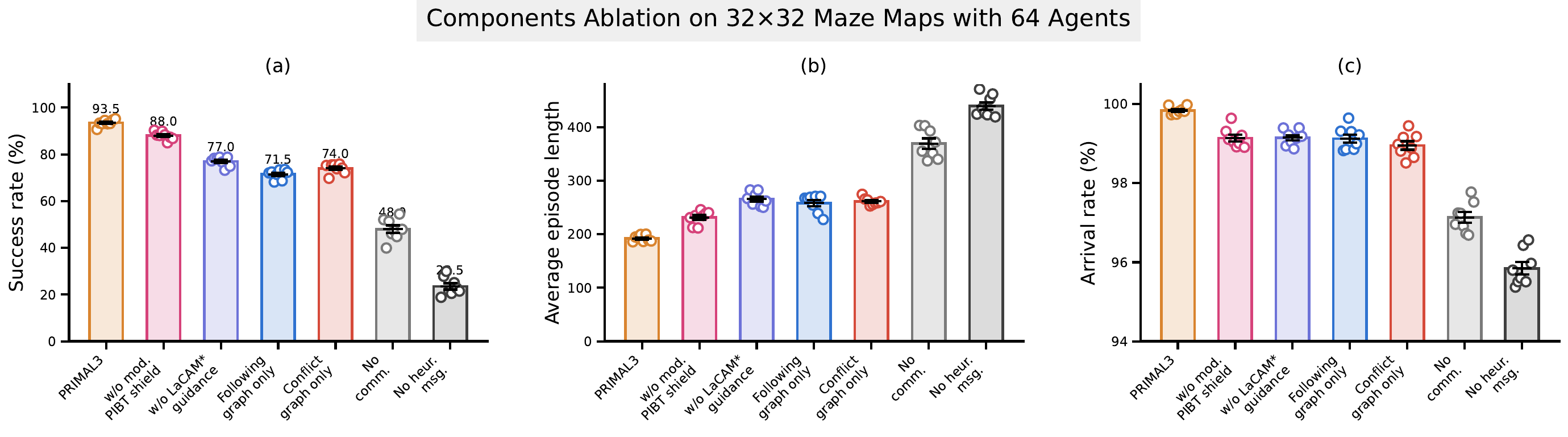}
    \caption{
    Components ablation on $32\times 32$ maze maps with 64 agents. 
    Panels show (a) success rate, (b) average episode length, and (c) arrival rate. 
    Numbers above the bars in panel (a) indicate success rates in percentage. 
    Bars denote the mean performance, open circles denote individual evaluation runs, and black error bars indicate mean $\pm$ SEM.
    }
    \label{fig:components_ablation}
\end{figure*}

We conduct component ablations on $32\times 32$ maze maps with 64 agents to evaluate the contribution of the proposed modified PIBT shielding mechanism, LaCAM* guidance, dual-graph communication, and heuristic features. 
The results are shown in Fig~\ref{fig:components_ablation}. 
Since a successful episode requires all agents to reach their goals, we use success rate as the primary metric, while average episode length and arrival rate provide complementary measurements of efficiency and individual goal-reaching behavior.

As shown in Fig~\ref{fig:components_ablation}(a), the full PRIMAL3 model achieves the highest success rate of $93.5\%$, together with the shortest average episode length in Fig~\ref{fig:components_ablation}(b). 
Replacing the modified PIBT shielding with the original PIBT shielding reduces the success rate from $93.5\%$ to $88.0\%$, indicating that the modified shielding provides more effective protection against invalid or poorly coordinated actions in dense maze environments. 
When LaCAM3 expert guidance is removed and the policy is trained purely with reinforcement learning, the success rate further drops to $77.0\%$, suggesting that expert-guided learning helps the policy acquire more reliable coordination behaviors.

We further isolate the effect of the dual-graph communication module under pure reinforcement learning. 
Using both the conflict and following graphs achieves a success rate of $77.0\%$, while using only the following graph or only the conflict graph obtains $71.5\%$ and $74.0\%$, respectively. 
This result shows that the two graphs capture complementary interaction patterns. 
The conflict graph is slightly more important in this dense maze setting, where head-on encounters, path blocking, and narrow-corridor conflicts are frequent. 
However, combining conflict-aware and following-aware communication still gives the best success rate, showing that same-direction coordination also contributes to stable multi-agent motion.

Removing both communication graphs causes a substantial degradation, reducing the success rate to $48.0\%$ and increasing the average episode length to $369.2$. 
This confirms that inter-agent communication is critical for resolving path overlaps and avoiding long-term congestion. 
When heuristic features are also removed, the success rate further decreases to $23.5\%$, and the average episode length increases to $439.1$. 
This indicates that heuristic path information provides an important prior for navigation, especially when agents cannot exchange information through the communication graphs.

Finally, Fig~\ref{fig:components_ablation}(c) shows that the arrival rate remains relatively high for most variants, even when the success rate drops noticeably. 
This suggests that many agents can still reach their goals individually, but solving the full multi-agent instance requires stronger coordination among all agents. 
Therefore, the success-rate degradation in Fig~\ref{fig:components_ablation}(a) more clearly reflects the importance of each component for multi-agent path finding.

\subsubsection{Decay Factors Analysis}
\label{sec:decay_factor_analysis}

\begin{figure}[t]
    \centering
    \includegraphics[width=\linewidth]{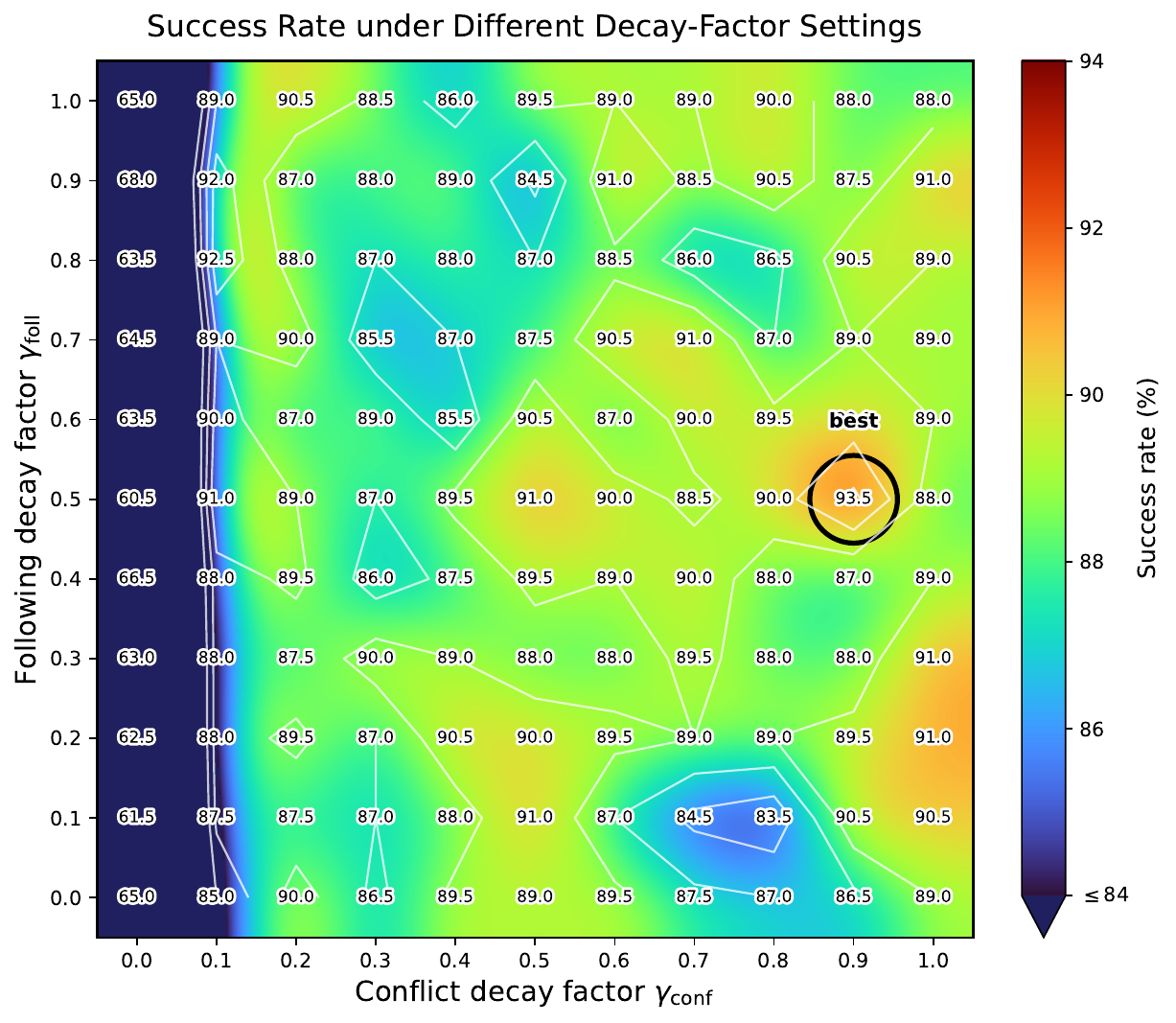}
    \caption{
    Success-rate landscape under different decay-factor settings on $32\times 32$ maze maps with 64 agents. 
    The horizontal axis denotes the conflict decay factor $\gamma_{\mathrm{conf}}$, and the vertical axis denotes the following decay factor $\gamma_{\mathrm{foll}}$. 
    The color field is smoothly interpolated for visualization, while the numbers indicate the success rates measured at the evaluated parameter settings. 
    The best result is obtained at $(\gamma_{\mathrm{conf}}, \gamma_{\mathrm{foll}})=(0.9,0.5)$.
    }
    \label{fig:decay_factor_heatmap}
\end{figure}

We further analyze the effect of the temporal decay factors used in the dual-graph construction. 
Specifically, we vary the conflict decay factor $\gamma_{\mathrm{conf}}$ and the following decay factor $\gamma_{\mathrm{foll}}$ from $0$ to $1$ with a step size of $0.1$ on $32\times 32$ maze maps with 64 agents. 
All other components are kept unchanged. 
Figure~\ref{fig:decay_factor_heatmap} shows the resulting success-rate landscape, where the horizontal axis corresponds to $\gamma_{\mathrm{conf}}$ and the vertical axis corresponds to $\gamma_{\mathrm{foll}}$.

The most prominent observation is that the conflict decay factor is essential. 
When $\gamma_{\mathrm{conf}}=0$, the success rate remains low across all values of $\gamma_{\mathrm{foll}}$, ranging only from $60.5\%$ to $68.0\%$. 
This indicates that using only instantaneous conflict overlap is insufficient in dense maze environments. 
Without temporal propagation of conflict information, agents cannot reliably anticipate future head-on encounters, blocking situations, or narrow-corridor interference.

Once $\gamma_{\mathrm{conf}}$ becomes positive, the success rate improves substantially. 
Most settings with $\gamma_{\mathrm{conf}}>0$ achieve success rates around $87\%\sim91\%$, showing that the model is relatively robust as long as future conflict information is preserved to some extent. 
The best performance, $93.5\%$, is achieved when $\gamma_{\mathrm{conf}}=0.9$ and $\gamma_{\mathrm{foll}}=0.5$. 
This supports the design choice that conflict interactions should be propagated over a relatively long temporal horizon, allowing agents to react early to potential blocking and collision-prone path overlaps.

In contrast, the following decay factor has a milder effect. 
For most nonzero values of $\gamma_{\mathrm{conf}}$, changing $\gamma_{\mathrm{foll}}$ does not cause as drastic a degradation as setting $\gamma_{\mathrm{conf}}$ to zero. 
This is consistent with the role of the following graph: same-direction overlaps mainly provide local consistency cues, such as maintaining coordinated motion along shared paths, rather than requiring long-range anticipation of future conflicts. 
A moderate following decay, such as $\gamma_{\mathrm{foll}}=0.5$, provides enough temporal context while avoiding excessive emphasis on distant same-direction overlaps.

Overall, the decay-factor analysis shows that the two factors play different roles. 
The conflict decay factor determines how far ahead agents consider future interference and is therefore critical for robust coordination in maze-like environments. 
The following decay factor mainly controls the temporal range of same-direction consistency cues and has a smaller but still useful influence. 
Based on this analysis, we use $(\gamma_{\mathrm{conf}}, \gamma_{\mathrm{foll}})=(0.9,0.5)$ in the final model.

\subsection{Real-Robot Experiments}
\label{sec:real_robot}

\begin{figure*}[t]
    \centering
    \includegraphics[width=\linewidth]{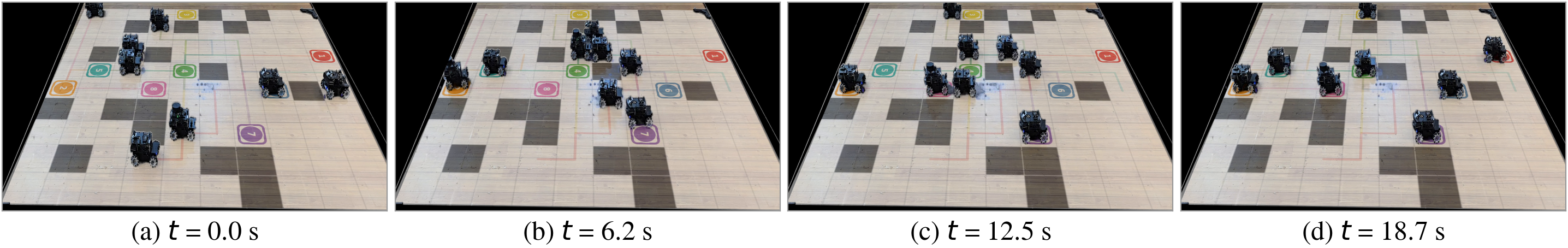}
    \vspace{0.2em}
    \includegraphics[width=\linewidth]{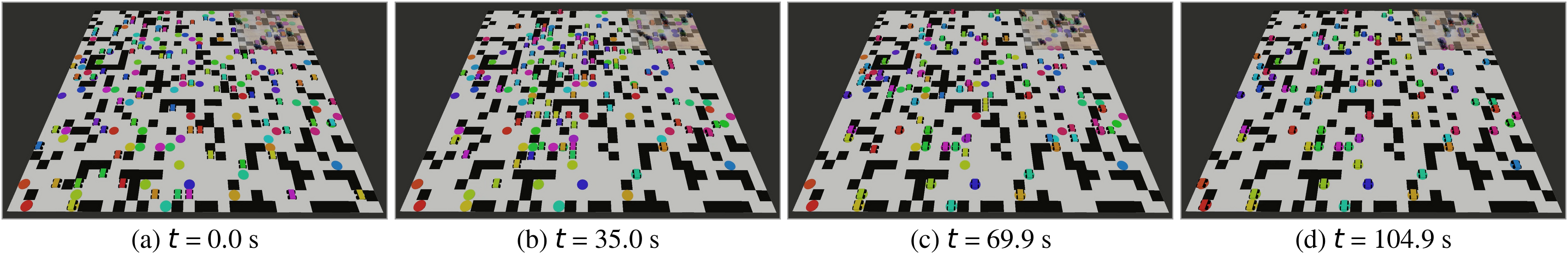}
    \caption{
    Real-robot evaluation of PRIMAL3.
    Panels show (above) 8 physical robots performing MAPF on a $10\times10$ random map and
    (bottom) a mixed-reality experiment with 8 physical robots and 92 virtual agents on a $32\times32$ random map.
    Each panel contains four frames.
    In bottom panel, the orange region in the upper-right corner denotes the physical workspace available to the real robots.
    Virtual agents may enter this region, whereas the physical robots remain within it due to the limited experimental space.
    }
    \label{fig:real_robot}
\end{figure*}

We conduct two proof-of-concept real-robot experiments to demonstrate the deployability of PRIMAL3. 
The first experiment involves eight physical robots performing MAPF on a $10\times10$ random map. 
The second considers a $32\times32$ map with 100 agents, comprising eight physical robots and 92 virtual agents. 
Although only eight agents are physically instantiated in the second experiment, PRIMAL3 jointly plans for all 100 agents; therefore, the physical robots must coordinate with both physical and virtual agents when resolving conflicts and selecting actions.

As PRIMAL3 operates on discrete grid states and outputs high-level grid actions, its policy does not depend on simulated robot dynamics. 
The planned actions can be directly deployed using a low-level controller that executes movements between adjacent grid cells; in our real-robot experiments, we employ P3GASUS~\cite{duhan2025p3gasus} for this purpose.
As shown in Fig~\ref{fig:real_robot}, both experiments were successfully executed, demonstrating the feasibility of transferring PRIMAL3 from simulation to physical multi-robot systems.

\section{Limitation and Future Work}

Despite these empirical improvements, PRIMAL3 has two main limitations. 
First, the current representation still relies on manually designed high-level features, such as cut vertices, dead-end regions, and blocking estimates. 
These features provide useful structural information over a spatial range that is difficult to recover directly from low-level local observations. 
However, their construction requires explicit access to the map topology and prior knowledge of which structural properties are relevant to coordination. 
Efficiently extracting scalable, structured representations of large maps directly from raw observations therefore remains an open problem.
Second, our current use of LaCAM3 does not capture the long-horizon inner/inter-dependence of agents' solutions. 
Although online expert intervention provides effective action-level guidance, the resulting supervision primarily treats expert actions as individual targets. 
It does not explicitly encode the temporal and inter-agent dependencies underlying the complete joint plan produced by LaCAM3. 
Moreover, repeatedly invoking LaCAM3 on states encountered during training introduces substantial computational overhead, making the overall training procedure time-consuming.

Future work will investigate scalable representation-learning methods that recover large-scale topological information directly from raw observations. 
An alternative direction is to learn a mapping from low-level observations to the manually designed structural features through auxiliary prediction or teacher-student distillation. 
In this formulation, the engineered features could serve as privileged supervision during training while no longer being required during execution. 
In addition, recent advances in robot imitation learning motivate the use of trajectory-level and more sample-efficient imitation methods. 
Distilling complete expert trajectories, modeling their long-horizon dependencies, and reusing expert experience through offline datasets could improve coordination while substantially reducing the number of online LaCAM3 queries required during training.

\section{Conclusion}
\label{sec:conclusion}

This paper introduced PRIMAL3, a topology-aware learning framework for multi-agent path finding. 
PRIMAL3 combines explicit structural node features, relation-specific dual-graph communication, LaCAM3-guided confidence boosting, and priority-aware PIBT action refinement. 
The following branch propagates multihop context among agents with compatible reference paths, whereas the conflict branch preserves relative information needed for differentiated decisions among agents competing for shared space. 
Topology-aware features further expose cut vertices, dead-end regions, remaining goal distances, and potential blocking effects. 
During training, LaCAM3 provides selective interventions and label-smoothed imitation targets for uncertain decisions. 
During execution, the PIBT module combines persistent, learned, and distance-aware agent priorities with policy-aware fallback preferences to produce collision-free one-step actions. 
Experiments demonstrate consistent improvements over existing learning-based baselines, including scalability to instances with up to $10{,}000$ agents, while the ablation results validate the contribution of each major component.

\bibliographystyle{IEEEtran}
\bibliography{ref}

@article{he2025social,
  title={Social behavior as a key to learning-based multi-agent pathfinding dilemmas},
  author={He, Chengyang and Duhan, Tanishq and Tulsyan, Parth and Kim, Patrick and Sartoretti, Guillaume},
  journal={Artificial Intelligence},
  pages={104397},
  year={2025},
  publisher={Elsevier}
}

@inproceedings{he2024alpha,
  title={Alpha: Attention-based long-horizon pathfinding in highly-structured areas},
  author={He, Chengyang and Yang, Tianze and Duhan, Tanishq and Wang, Yutong and Sartoretti, Guillaume},
  booktitle={2024 IEEE International Conference on Robotics and Automation (ICRA)},
  pages={14576--14582},
  year={2024},
  organization={IEEE}
}

@inproceedings{stern2019multi,
  title={Multi-agent pathfinding: Definitions, variants, and benchmarks},
  author={Stern, Roni and Sturtevant, Nathan and Felner, Ariel and Koenig, Sven and Ma, Hang and Walker, Thayne and Li, Jiaoyang and Atzmon, Dor and Cohen, Liron and Kumar, TK and others},
  booktitle={Proceedings of the international symposium on combinatorial search},
  volume={10},
  number={1},
  pages={151--158},
  year={2019}
}

@article{sartoretti2019primal,
  title={Primal: Pathfinding via reinforcement and imitation multi-agent learning},
  author={Sartoretti, Guillaume and Kerr, Justin and Shi, Yunfei and Wagner, Glenn and Kumar, TK Satish and Koenig, Sven and Choset, Howie},
  journal={IEEE Robotics and Automation Letters},
  volume={4},
  number={3},
  pages={2378--2385},
  year={2019},
  publisher={IEEE}
}

@article{wang2020mobile,
  title={Mobile robot path planning in dynamic environments through globally guided reinforcement learning},
  author={Wang, Binyu and Liu, Zhe and Li, Qingbiao and Prorok, Amanda},
  journal={IEEE Robotics and Automation Letters},
  volume={5},
  number={4},
  pages={6932--6939},
  year={2020},
  publisher={IEEE}
}

@inproceedings{liu2020mapper,
  title={Mapper: Multi-agent path planning with evolutionary reinforcement learning in mixed dynamic environments},
  author={Liu, Zuxin and Chen, Baiming and Zhou, Hongyi and Koushik, Guru and Hebert, Martial and Zhao, Ding},
  booktitle={2020 IEEE/RSJ International Conference on Intelligent Robots and Systems (IROS)},
  pages={11748--11754},
  year={2020},
  organization={IEEE}
}

@inproceedings{ma2021distributed,
  title={Distributed heuristic multi-agent path finding with communication},
  author={Ma, Ziyuan and Luo, Yudong and Ma, Hang},
  booktitle={2021 IEEE International Conference on Robotics and Automation (ICRA)},
  pages={8699--8705},
  year={2021},
  organization={IEEE}
}

@inproceedings{li2022multi,
  title={Multi-agent path finding with prioritized communication learning},
  author={Li, Wenhao and Chen, Hongjun and Jin, Bo and Tan, Wenzhe and Zha, Hongyuan and Wang, Xiangfeng},
  booktitle={2022 International Conference on Robotics and Automation (ICRA)},
  pages={10695--10701},
  year={2022},
  organization={IEEE}
}

@inproceedings{wang2023scrimp,
  title={Scrimp: Scalable communication for reinforcement-and imitation-learning-based multi-agent pathfinding},
  author={Wang, Yutong and Xiang, Bairan and Huang, Shinan and Sartoretti, Guillaume},
  booktitle={2023 IEEE/RSJ International Conference on Intelligent Robots and Systems (IROS)},
  pages={9301--9308},
  year={2023},
  organization={IEEE}
}

@article{ma2021learning,
  title={Learning selective communication for multi-agent path finding},
  author={Ma, Ziyuan and Luo, Yudong and Pan, Jia},
  journal={IEEE Robotics and Automation Letters},
  volume={7},
  number={2},
  pages={1455--1462},
  year={2021},
  publisher={IEEE}
}

@inproceedings{liao2025sigma,
  title={Sigma: Sheaf-informed geometric multi-agent pathfinding},
  author={Liao, Shuhao and Xia, Weihang and Cao, Yuhong and Dai, Weiheng and He, Chengyang and Wu, Wenjun and Sartoretti, Guillaume},
  booktitle={2025 IEEE International Conference on Robotics and Automation (ICRA)},
  pages={1--7},
  year={2025},
  organization={IEEE}
}

@inproceedings{li2020graph,
  title={Graph neural networks for decentralized multi-robot path planning},
  author={Li, Qingbiao and Gama, Fernando and Ribeiro, Alejandro and Prorok, Amanda},
  booktitle={2020 IEEE/RSJ international conference on intelligent robots and systems (IROS)},
  pages={11785--11792},
  year={2020},
  organization={IEEE}
}

@article{li2021message,
  title={Message-aware graph attention networks for large-scale multi-robot path planning},
  author={Li, Qingbiao and Lin, Weizhe and Liu, Zhe and Prorok, Amanda},
  journal={IEEE Robotics and Automation Letters},
  volume={6},
  number={3},
  pages={5533--5540},
  year={2021},
  publisher={IEEE}
}

@inproceedings{veerapaneni2024improving,
  title={Improving learnt local MAPF policies with heuristic search},
  author={Veerapaneni, Rishi and Wang, Qian and Ren, Kevin and Jakobsson, Arthur and Li, Jiaoyang and Likhachev, Maxim},
  booktitle={Proceedings of the International Conference on Automated Planning and Scheduling},
  volume={34},
  pages={597--606},
  year={2024}
}

@inproceedings{veerapaneni2025work,
  title={Work Smarter Not Harder: Simple Imitation Learning with CS-PIBT Outperforms Large-Scale Imitation Learning for MAPF},
  author={Veerapaneni, Rishi and Jakobsson, Arthur and Ren, Kevin and Kim, Samuel and Li, Jiaoyang and Likhachev, Maxim},
  booktitle={2025 IEEE International Conference on Robotics and Automation (ICRA)},
  pages={10229--10236},
  year={2025},
  organization={IEEE}
}

@article{virmani2021subdimensional,
  title={Subdimensional expansion using attention-based learning for multi-agent path finding},
  author={Virmani, Lakshay and Ren, Zhongqiang and Rathinam, Sivakumar and Choset, Howie},
  journal={arXiv preprint arXiv:2109.14695},
  year={2021}
}

@inproceedings{morag2023adapting,
  title={Adapting to Planning Failures in Lifelong Multi-Agent Path Finding},
  author={Morag, Jonathan and Stern, Roni and Felner, Ariel},
  booktitle={Proceedings of the International Symposium on Combinatorial Search},
  volume={16},
  number={1},
  pages={47--55},
  year={2023}
}

@inproceedings{andreychuk2025mapf,
  title={Mapf-gpt: Imitation learning for multi-agent pathfinding at scale},
  author={Andreychuk, Anton and Yakovlev, Konstantin and Panov, Aleksandr and Skrynnik, Alexey},
  booktitle={Proceedings of the AAAI Conference on Artificial Intelligence},
  volume={39},
  number={22},
  pages={23126--23134},
  year={2025}
}

@inproceedings{andreychuk2025advancing,
  title={Advancing Learnable Multi-Agent Pathfinding Solvers with Active Fine-Tuning},
  author={Andreychuk, Anton and Yakovlev, Konstantin and Panov, Aleksandr and Skrynnik, Alexey},
  booktitle={2025 IEEE/RSJ International Conference on Intelligent Robots and Systems (IROS)},
  pages={10564--10571},
  year={2025},
  organization={IEEE}
}

@article{jain2026pairwise,
  title={Pairwise is Not Enough: Hypergraph Neural Networks for Multi-Agent Pathfinding},
  author={Jain, Rishabh and Okumura, Keisuke and Amir, Michael and Lio, Pietro and Prorok, Amanda},
  journal={arXiv preprint arXiv:2602.06733},
  year={2026}
}

@article{gao2023review,
  title={A review of graph-based multi-agent pathfinding solvers: From classical to beyond classical},
  author={Gao, Jianqi and Li, Yanjie and Li, Xinyi and Yan, Kejian and Lin, Ke and Wu, Xinyu},
  journal={Knowledge-Based Systems},
  pages={111121},
  year={2023},
  publisher={Elsevier}
}

@inproceedings{wagner2011m,
  title={M*: A complete multirobot path planning algorithm with performance bounds},
  author={Wagner, Glenn and Choset, Howie},
  booktitle={2011 IEEE/RSJ international conference on intelligent robots and systems},
  pages={3260--3267},
  year={2011},
  organization={IEEE}
}

@article{sharon2015conflict,
  title={Conflict-based search for optimal multi-agent pathfinding},
  author={Sharon, Guni and Stern, Roni and Felner, Ariel and Sturtevant, Nathan R},
  journal={Artificial intelligence},
  volume={219},
  pages={40--66},
  year={2015},
  publisher={Elsevier}
}

@inproceedings{boyarski2015icbs,
  title={Icbs: The improved conflict-based search algorithm for multi-agent pathfinding},
  author={Boyarski, Eli and Felner, Ariel and Stern, Roni and Sharon, Guni and Betzalel, Oded and Tolpin, David and Shimony, Eyal},
  booktitle={Proceedings of the International Symposium on Combinatorial Search},
  volume={6},
  number={1},
  pages={223--225},
  year={2015}
}

@inproceedings{felner2018adding,
  title={Adding heuristics to conflict-based search for multi-agent path finding},
  author={Felner, Ariel and Li, Jiaoyang and Boyarski, Eli and Ma, Hang and Cohen, Liron and Kumar, TK Satish and Koenig, Sven},
  booktitle={Proceedings of the International Conference on Automated Planning and Scheduling},
  volume={28},
  pages={83--87},
  year={2018}
}

@inproceedings{li2019improved,
  title={Improved Heuristics for Multi-Agent Path Finding with Conflict-Based Search.},
  author={Li, Jiaoyang and Felner, Ariel and Boyarski, Eli and Ma, Hang and Koenig, Sven},
  booktitle={IJCAI},
  volume={2019},
  pages={442--449},
  year={2019}
}

@inproceedings{li2019symmetry,
  title={Symmetry-breaking constraints for grid-based multi-agent path finding},
  author={Li, Jiaoyang and Harabor, Daniel and Stuckey, Peter J and Ma, Hang and Koenig, Sven},
  booktitle={Proceedings of the AAAI conference on artificial intelligence},
  volume={33},
  number={01},
  pages={6087--6095},
  year={2019}
}

@inproceedings{li2020new,
  title={New techniques for pairwise symmetry breaking in multi-agent path finding},
  author={Li, Jiaoyang and Gange, Graeme and Harabor, Daniel and Stuckey, Peter J and Ma, Hang and Koenig, Sven},
  booktitle={Proceedings of the International Conference on Automated Planning and Scheduling},
  volume={30},
  pages={193--201},
  year={2020}
}

@article{li2021pairwise,
  title={Pairwise symmetry reasoning for multi-agent path finding search},
  author={Li, Jiaoyang and Harabor, Daniel and Stuckey, Peter J and Ma, Hang and Gange, Graeme and Koenig, Sven},
  journal={Artificial Intelligence},
  volume={301},
  pages={103574},
  year={2021},
  publisher={Elsevier}
}

@inproceedings{li2019disjoint,
  title={Disjoint splitting for multi-agent path finding with conflict-based search},
  author={Li, Jiaoyang and Harabor, Daniel and Stuckey, Peter J and Felner, Ariel and Ma, Hang and Koenig, Sven},
  booktitle={Proceedings of the international conference on automated planning and scheduling},
  volume={29},
  pages={279--283},
  year={2019}
}

@article{wagner2015subdimensional,
  title={Subdimensional expansion for multirobot path planning},
  author={Wagner, Glenn and Choset, Howie},
  journal={Artificial intelligence},
  volume={219},
  pages={1--24},
  year={2015},
  publisher={Elsevier}
}

@inproceedings{barer2014suboptimal,
  title={Suboptimal variants of the conflict-based search algorithm for the multi-agent pathfinding problem},
  author={Barer, Max and Sharon, Guni and Stern, Roni and Felner, Ariel},
  booktitle={Proceedings of the international symposium on combinatorial Search},
  volume={5},
  number={1},
  pages={19--27},
  year={2014}
}

@inproceedings{li2021eecbs,
  title={Eecbs: A bounded-suboptimal search for multi-agent path finding},
  author={Li, Jiaoyang and Ruml, Wheeler and Koenig, Sven},
  booktitle={Proceedings of the AAAI conference on artificial intelligence},
  volume={35},
  number={14},
  pages={12353--12362},
  year={2021}
}

@article{zhang2022multi,
  title={Multi-agent path finding with mutex propagation},
  author={Zhang, Han and Li, Jiaoyang and Surynek, Pavel and Kumar, TK Satish and Koenig, Sven},
  journal={Artificial Intelligence},
  volume={311},
  pages={103766},
  year={2022},
  publisher={Elsevier}
}

@article{li2025multi,
  title={Multi-Agent Path Finding via Finite-Horizon Hierarchical Factorization},
  author={Li, Jiarui and Zanardi, Alessandro and Zardini, Gioele},
  journal={arXiv preprint arXiv:2505.07779},
  year={2025}
}

@article{li2025fico,
  title={FICO: Finite-Horizon Closed-Loop Factorization for Unified Multi-Agent Path Finding},
  author={Li, Jiarui and Zanardi, Alessandro and Pecora, Federico and Zhang, Runyu and Zardini, Gioele},
  journal={arXiv preprint arXiv:2511.13961},
  year={2025}
}

@article{li2026adaptive,
  title={Adaptive-Horizon Conflict-Based Search for Closed-Loop Multi-Agent Path Finding},
  author={Li, Jiarui and Pecora, Federico and Zhang, Runyu and Zardini, Gioele},
  journal={arXiv preprint arXiv:2602.12024},
  year={2026}
}

@inproceedings{li2022mapf,
  title={MAPF-LNS2: Fast repairing for multi-agent path finding via large neighborhood search},
  author={Li, Jiaoyang and Chen, Zhe and Harabor, Daniel and Stuckey, Peter J and Koenig, Sven},
  booktitle={Proceedings of the AAAI Conference on Artificial Intelligence},
  volume={36},
  number={9},
  pages={10256--10265},
  year={2022}
}

@inproceedings{li2021anytime,
  title={Anytime multi-agent path finding via large neighborhood search},
  author={Li, Jiaoyang and Chen, Zhe and Harabor, Daniel and Stuckey, Peter J and Koenig, Sven},
  booktitle={International joint conference on artificial intelligence 2021},
  pages={4127--4135},
  year={2021},
  organization={Association for the Advancement of Artificial Intelligence (AAAI)}
}

@inproceedings{silver2005cooperative,
  title={Cooperative pathfinding},
  author={Silver, David},
  booktitle={Proceedings of the aaai conference on artificial intelligence and interactive digital entertainment},
  volume={1},
  number={1},
  pages={117--122},
  year={2005}
}

@article{okumura2023engineering,
  title={Engineering LaCAM*: Towards Real-Time, Large-Scale, and Near-Optimal Multi-Agent Pathfinding},
  author={Okumura, Keisuke},
  journal={arXiv preprint arXiv:2308.04292},
  year={2023}
}

@inproceedings{okumura2023lacam,
  title={Lacam: Search-based algorithm for quick multi-agent pathfinding},
  author={Okumura, Keisuke},
  booktitle={Proceedings of the AAAI Conference on Artificial Intelligence},
  volume={37},
  number={10},
  pages={11655--11662},
  year={2023}
}

@article{okumura2022priority,
  title={Priority inheritance with backtracking for iterative multi-agent path finding},
  author={Okumura, Keisuke and Machida, Manao and D{\'e}fago, Xavier and Tamura, Yasumasa},
  journal={Artificial Intelligence},
  volume={310},
  pages={103752},
  year={2022},
  publisher={Elsevier}
}

@inproceedings{ma2019lifelong,
  title={Lifelong path planning with kinematic constraints for multi-agent pickup and delivery},
  author={Ma, Hang and H{\"o}nig, Wolfgang and Kumar, TK Satish and Ayanian, Nora and Koenig, Sven},
  booktitle={Proceedings of the AAAI Conference on Artificial Intelligence},
  volume={33},
  number={01},
  pages={7651--7658},
  year={2019}
}

@article{skrynnik2023learn,
  title={Learn to Follow: Decentralized Lifelong Multi-agent Pathfinding via Planning and Learning},
  author={Skrynnik, Alexey and Andreychuk, Anton and Nesterova, Maria and Yakovlev, Konstantin and Panov, Aleksandr},
  journal={arXiv preprint arXiv:2310.01207},
  year={2023}
}

@article{damani2021primal,
  title={PRIMAL $ \_2 $: Pathfinding via reinforcement and imitation multi-agent learning-lifelong},
  author={Damani, Mehul and Luo, Zhiyao and Wenzel, Emerson and Sartoretti, Guillaume},
  journal={IEEE Robotics and Automation Letters},
  volume={6},
  number={2},
  pages={2666--2673},
  year={2021},
  publisher={IEEE}
}

@article{chandra2023socialmapf,
  title={Socialmapf: Optimal and efficient multi-agent path finding with strategic agents for social navigation},
  author={Chandra, Rohan and Maligi, Rahul and Anantula, Arya and Biswas, Joydeep},
  journal={IEEE Robotics and Automation Letters},
  year={2023},
  publisher={IEEE}
}

@article{battaglia2018relational,
  title={Relational inductive biases, deep learning, and graph networks},
  author={Battaglia, Peter W and Hamrick, Jessica B and Bapst, Victor and Sanchez-Gonzalez, Alvaro and Zambaldi, Vinicius and Malinowski, Mateusz and Tacchetti, Andrea and Raposo, David and Santoro, Adam and Faulkner, Ryan and others},
  journal={arXiv preprint arXiv:1806.01261},
  year={2018}
}

@article{chen2022nagphormer,
  title={NAGphormer: A tokenized graph transformer for node classification in large graphs},
  author={Chen, Jinsong and Gao, Kaiyuan and Li, Gaichao and He, Kun},
  journal={arXiv preprint arXiv:2206.04910},
  year={2022}
}

@inproceedings{li2021lifelong,
  title={Lifelong multi-agent path finding in large-scale warehouses},
  author={Li, Jiaoyang and Tinka, Andrew and Kiesel, Scott and Durham, Joseph W and Kumar, TK Satish and Koenig, Sven},
  booktitle={Proceedings of the AAAI Conference on Artificial Intelligence},
  volume={35},
  number={13},
  pages={11272--11281},
  year={2021}
}

@article{duhan2025p3gasus,
  title={P3GASUS: Pre-Planned Path Execution Graphs for Multi-Agent Systems at Ultra-Large Scale},
  author={Duhan, Tanishq and He, Chengyang and Sartoretti, Guillaume},
  journal={IEEE Robotics and Automation Letters},
  volume={11},
  number={2},
  pages={1274--1281},
  year={2025},
  publisher={IEEE}
}

@inproceedings{li2023intersection,
  title={Intersection coordination with priority-based search for autonomous vehicles},
  author={Li, Jiaoyang and Lin, Eugene and Vu, Hai L and Koenig, Sven and others},
  booktitle={Proceedings of the AAAI Conference on Artificial Intelligence},
  volume={37},
  number={10},
  pages={11578--11585},
  year={2023}
}

@inproceedings{ma2017feasibility,
  title={Feasibility study: Moving non-homogeneous teams in congested video game environments},
  author={Ma, Hang and Yang, Jingxing and Cohen, Liron and Kumar, TK and Koenig, Sven},
  booktitle={Proceedings of the AAAI Conference on Artificial Intelligence and Interactive Digital Entertainment},
  volume={13},
  number={1},
  pages={270--272},
  year={2017}
}

@inproceedings{jiang2025deploying,
  title={Deploying ten thousand robots: Scalable imitation learning for lifelong multi-agent path finding},
  author={Jiang, He and Wang, Yutong and Veerapaneni, Rishi and Duhan, Tanishq and Sartoretti, Guillaume and Li, Jiaoyang},
  booktitle={2025 IEEE International Conference on Robotics and Automation (ICRA)},
  pages={1--7},
  year={2025},
  organization={IEEE}
}

@inproceedings{zang2025online,
  title={Online guidance graph optimization for lifelong multi-agent path finding},
  author={Zang, Hongzhi and Zhang, Yulun and Jiang, He and Chen, Zhe and Harabor, Daniel and Stuckey, Peter J and Li, Jiaoyang},
  booktitle={Proceedings of the AAAI Conference on Artificial Intelligence},
  volume={39},
  number={14},
  pages={14726--14735},
  year={2025}
}

@article{zheng2026learning,
  title={Learning-guided prioritized planning for lifelong multi-agent path finding in warehouse automation},
  author={Zheng, Han and Ma, Yining and Araki, Brandon and Chen, Jingkai and Wu, Cathy},
  journal={Journal of Artificial Intelligence Research},
  volume={85},
  year={2026}
}

@article{zhang2026optimization,
  title={Optimization of Edge Directions and Weights for Mixed Guidance Graphs in Lifelong Multi-Agent Path Finding},
  author={Zhang, Yulun and Bhatt, Varun and Fontaine, Matthew C and Nikolaidis, Stefanos and Li, Jiaoyang},
  journal={arXiv preprint arXiv:2602.23468},
  year={2026}
}

@article{shaoul2025collaborative,
  title={Collaborative Multi-Robot Non-Prehensile Manipulation via Flow-Matching Co-Generation},
  author={Shaoul, Yorai and Chen, Zhe and Mohamed, Mohamed Naveed Gul and Pecora, Federico and Likhachev, Maxim and Li, Jiaoyang},
  journal={arXiv preprint arXiv:2511.10874},
  year={2025}
}

@article{shaoul2024multi,
  title={Multi-robot motion planning with diffusion models},
  author={Shaoul, Yorai and Mishani, Itamar and Vats, Shivam and Li, Jiaoyang and Likhachev, Maxim},
  journal={arXiv preprint arXiv:2410.03072},
  year={2024}
}

@article{zhang2025flow,
  title={Flow-Based Task Assignment for Large-Scale Online Multi-Agent Pickup and Delivery},
  author={Zhang, Yue and Chen, Zhe and Harabor, Daniel and Bodic, Pierre Le and Stuckey, Peter J},
  journal={arXiv preprint arXiv:2508.05890},
  year={2025}
}

@article{ma2017lifelong,
  title={Lifelong multi-agent path finding for online pickup and delivery tasks},
  author={Ma, Hang and Li, Jiaoyang and Kumar, TK and Koenig, Sven},
  journal={arXiv preprint arXiv:1705.10868},
  year={2017}
}

@article{nelson1972theory,
  title={Theory and applications of hazard plotting for censored failure data},
  author={Nelson, Wayne},
  journal={Technometrics},
  volume={14},
  number={4},
  pages={945--966},
  year={1972},
  publisher={Taylor \& Francis}
}

@book{kalbfleisch2002statistical,
  title={The statistical analysis of failure time data},
  author={Kalbfleisch, John D and Prentice, Ross L},
  year={2002},
  publisher={John Wiley \& Sons}
}

\end{document}